\documentclass{article}

    \usepackage[preprint]{neurips_2024}

\usepackage[utf8]{inputenc} 
\usepackage[T1]{fontenc}    
\usepackage{url}            
\usepackage{booktabs}       
\usepackage{amsfonts}       
\usepackage{nicefrac}       
\usepackage{microtype}      
\usepackage[dvipsnames]{xcolor}

\usepackage{arydshln}

\usepackage{amsfonts}
\usepackage{mathtools}
\usepackage{natbib}
\setcitestyle{square}
\setcitestyle{numbers}
\usepackage{xcolor}
\usepackage{url}
\usepackage{amsmath}
\usepackage{amssymb}
\usepackage{amsthm}
\usepackage{graphicx}
\usepackage{subcaption}
\usepackage{nicefrac}
\usepackage{appendix}
\usepackage{enumitem}
\usepackage{arydshln}
\usepackage{adjustbox}
\definecolor{darker}{rgb}{0,0.15,0.7}
\definecolor{emerald}{rgb}{0, 0.52, 0.39}
\definecolor{black}{rgb}{0, 0, 0}
\usepackage[colorlinks, urlcolor=darker, citecolor=darker, linkcolor=black,linktoc=all]{hyperref} 

\usepackage{lastpage}
\usepackage{fancyhdr}

\usepackage{tikz}
\usetikzlibrary{lindenmayersystems}

\pgfdeclarelindenmayersystem{A}{
    \rule{F -> FF[+F][-F]}
}

\pgfdeclarelindenmayersystem{B}{
    \rule{F -> ffF[++FF][--FF]}
}

\pgfdeclarelindenmayersystem{C}{
    \symbol{G}{\pgflsystemdrawforward}
    \rule{F -> F[+F][-F]FG[+F][-F]FG}
}

\tikzset{
    type/.style={l-system={#1, axiom=F,order=3,step=4pt,angle=60},
      blue, opacity=0.4, line width=.5mm, line cap=round   
    },
}

\usepackage{multirow}
\usepackage{adjustbox}

\usepackage{lipsum}

\usepackage{amsmath}
\usepackage{amssymb}
\usepackage{mathtools}
\usepackage{amsthm}

\usepackage[inkscapelatex=false]{svg}

\newcommand\blfootnote[1]{%
  \begingroup
  \renewcommand\thefootnote{}\footnote{#1}%
  \addtocounter{footnote}{-1}%
  \endgroup
}

\title{NVLM: Open Frontier-Class Multimodal LLMs} 

\author{
\and{\hspace{-1cm}Wenliang Dai$^{*}$\qquad Nayeon Lee$^{*}$\qquad Boxin Wang$^{*}$\qquad Zhuolin Yang$^{*}$}
\vspace{0.15cm}
\and{Zihan Liu\ \ \ Jon Barker\ \ \ Tuomas Rintamaki\ \ \ Mohammad Shoeybi\ \ \ Bryan Catanzaro}
\vspace{0.15cm}
\and{Wei Ping$^{*, \dagger}$}
\vspace{0.15cm}
 \vspace{.2cm} \\ 
\textbf{NVIDIA} 
\vspace{1.9mm} \\
$^{*}$~Equal contributions, ordered alphabetically \\
\vspace{1mm}
\texttt{\{wdai, nayeonl, boxinw, zhuoliny, wping\}@nvidia.com} \\
\vspace{1mm}
$^{\dagger}$~Leads the effort.
}

\begin{document}


\blfootnote{$^{1}$~\textbf{N}VIDIA \textbf{Vi}sion \textbf{L}anguage \textbf{M}odel.}

\maketitle

\begin{abstract}
We introduce NVLM 1.0,~$^{1}$ a family of frontier-class multimodal large language models (LLMs) that achieve state-of-the-art results on vision-language tasks, rivaling the leading proprietary models (e.g., GPT-4o) and open-access models (e.g., Llama 3-V 405B and InternVL 2). Remarkably, NVLM 1.0 shows improved text-only performance over its LLM backbone after multimodal training.

In terms of \textbf{model design}, we perform a comprehensive comparison between decoder-only multimodal LLMs (e.g., LLaVA) and cross-attention-based models (e.g., Flamingo). Based on the strengths and weaknesses of both approaches, we propose a novel architecture that enhances both training efficiency and multimodal reasoning capabilities.
Furthermore, we introduce a 1-D tile-tagging design for tile-based dynamic high-resolution images, which significantly boosts performance on multimodal reasoning and OCR-related tasks.
Regarding \textbf{training data}, we meticulously curate and provide detailed information on our multimodal pretraining and supervised fine-tuning datasets. Our findings indicate that dataset quality and task diversity are more important than scale, even during the pretraining phase, across all architectures.
Notably, we develop \textbf{production-grade multimodality} for the NVLM-1.0 models, enabling them to excel in vision-language tasks while maintaining and even improving text-only performance compared to their LLM backbones. 
To achieve this, we craft and integrate a high-quality text-only dataset into multimodal training, alongside a substantial amount of multimodal math and reasoning data, leading to enhanced math and coding capabilities across modalities.

To advance research in the field, we release the model weights at \url{https://huggingface.co/nvidia/NVLM-D-72B} and will open-source the training code for the community soon. For more information, please visit our project page at \url{https://research.nvidia.com/labs/adlr/NVLM-1}.
\end{abstract}

\newpage

\pagenumbering{arabic}

\tableofcontents

\begin{figure}
\centerline{\includegraphics[width=1.01\linewidth]{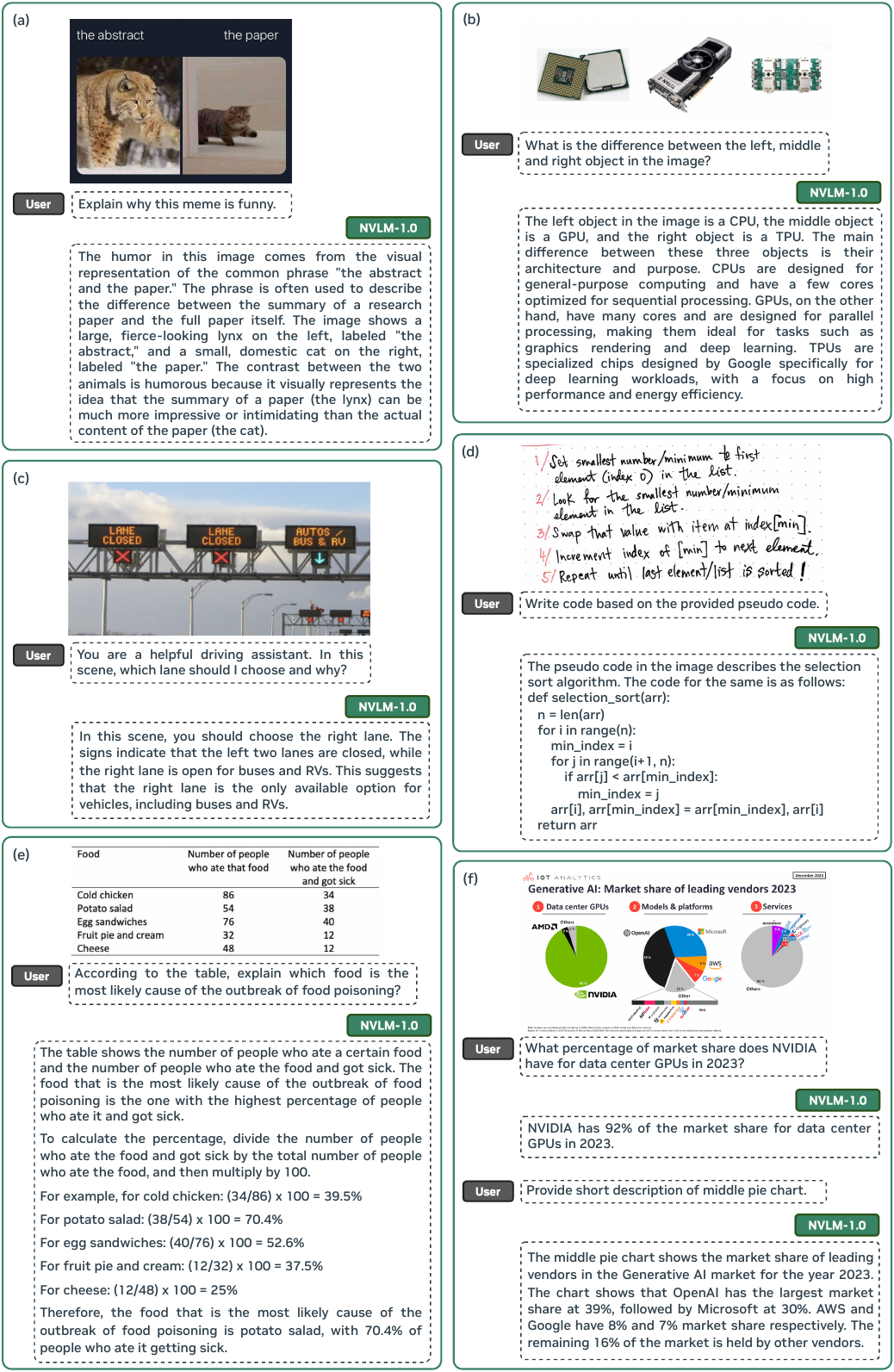}}
\caption{\footnotesize
Qualitative examples generated by our NVLM-D$_{ 1.0}$72B model. We demonstrate diverse capabilities of our model, including chart and table understanding, OCR, localization, knowledge-grounded image description, humorous meme understanding, scene understanding, math reasoning and coding capabilities. For more examples, refer to Appendix~\ref{appendix:more_examples}.
}
\label{fig:qualitative_example}
\end{figure}

\newpage

\section{Introduction}
\label{sec:intro}

\begin{figure}
\centerline{\includegraphics[width=1.01\linewidth]{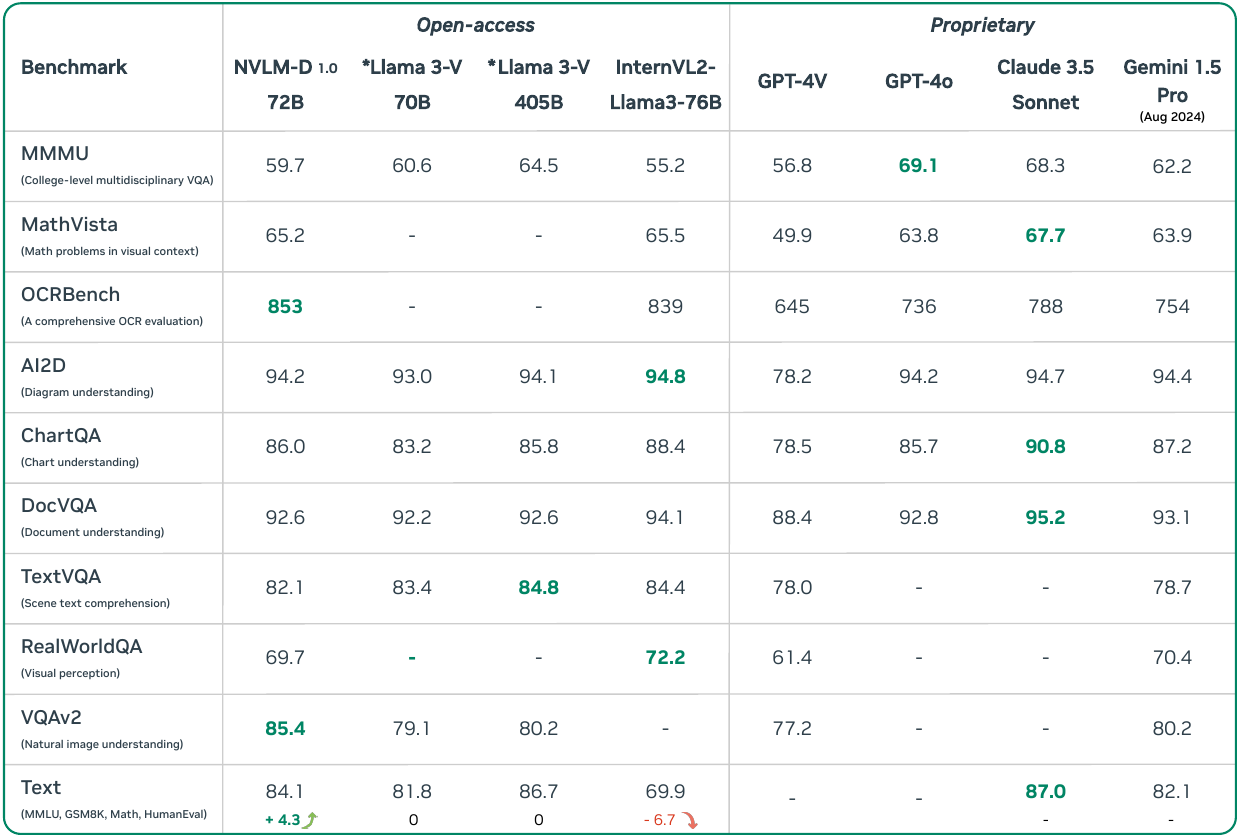}}
\vspace{-1mm}
\caption{\footnotesize
NVLM 1.0 \emph{versus} leading proprietary and open-access multimodal LLMs. Note that the model weights for $^*$Llama 3-V~\citep{llama3-1} have not been released as of the time of this report.
The results demonstrate that NVLM~1.0 achieves performance on par with leading models across both vision-language and text-only tasks. 
Additionally, we compare multimodal LLM to its backbone LLM on text-only tasks. Llama 3-V 70B and 405B show no degradation in text-only tasks, as their LLM backbones are frozen during multimodal training. In contrast, our NVLM-D$_{ 1.0}$ 72B model demonstrates significant improvements over its text backbone on text-only math and coding benchmarks, with average accuracy increasing by 4.3 points after multimodal training.
See Table~\ref{tab:main_result} and Table~\ref{tab:text-only} for full details.
}
\label{fig:Table0}
\end{figure}

Large language models~(LLMs)~\citep{brown2020language} have laid the foundation for the rapid progress in AI recently.
Since the introduction of ChatGPT~\citep{chatgpt}, LLMs have revolutionized the text domain and are becoming universal task solvers for natural language processing, math and coding problems. 
Simultaneously, multimodal LLMs~(MLLMs)~\citep{alayrac2022flamingo, gpt4v}, which bridge the physical world with language models, have gained significant traction.
The release of GPT-4V~\citep{gpt4v} has sparked a competitive race in the development of proprietary multimodal LLMs for vision-language intelligence~\citep{gemini1-0, gpt4o, claude3, claude3.5sonnet, xai2024Grok1-5, xai2024Grok2, reka-core}.
However, the model architectures, training data, and methods used to build these proprietary models remain undisclosed, preventing the research community from building upon them.

A notable feature of leading proprietary multimodal LLMs is their exceptional performance on both multimodal and text-only tasks, a quality we refer to as \emph{production-grade multimodality}~\citep{gpt4o, gemini1-0, gemini1-5}.
For example, GPT-4o is a single neural network trained end-to-end on text and images, achieving state-of-the-art results in both text-only and vision-language tasks~\citep{gpt4osystem-card}.
This unified approach simplifies deployment by eliminating the need to route different input modalities to separate LLMs, offering users a seamless experience for switching between modalities without losing text or multimodal context.

The community has made significant progress in advancing the capabilities of open-access multimodal LLMs~\citep{dai2023instructblip, liu2023llava, chen2024internvl1-5, lin2024vila, tong2024cambrian}. Notable families of open models include BLIP~\citep{li2022blip, li2023blip, dai2023instructblip}, LLaVA~\citep{liu2023llava, liu2023improvedllava, liu2024llavanext, li2024llava-onevision}, InternVL~\citep{chen2024internvl, chen2024internvl1-5, 2024internvl2}, and Llama 3-V~\citep{llama3-1}.
The most common architectures used to build these multimodal LLMs are the decoder-only architecture (e.g., LLaVA~\citep{liu2023llava} and InternVL~\citep{chen2024internvl1-5}), which processes image tokens within the LLM self-attention layers, and the cross-attention-based architecture (e.g., Flamingo~\citep{alayrac2022flamingo} and Llama 3-V~\citep{llama3-1}), which handles image tokens through LLM cross-attention layers.

However, the previous studies of multimodal LLMs have several limitations:
\vspace{-0.4em}
\begin{enumerate}[leftmargin=1.7em]
\item[$\bullet$]
In contrast to the convergence of model architectures to build LLM in the text domain~\citep{brown2020language, smith2022using, gpt4,claude3,gemini1-0}, i.e., the decoder-only transformer~\citep{vaswani2017attention}, existing multimodal LLM architectures~(e.g., decoder-only vs. cross-attention models) have not been studied and compared in an apples-to-apples manner. There is no information regarding the architectures of proprietary models. Furthermore, studies on open-access models differ in their choice of LLM backbones, vision encoders, and, most importantly, training data, making direct comparisons challenging. For these reasons, IDEFICS-80B, an open-access reproduction of Flamingo~\citep{laurencon2023IDEFICS} based on LLaMA-65B~\citep{touvron2023llama}, is perceived as significantly lagging behind LLaVA-1.5-13B~\citep{liu2023improvedllava}, which is based on Vicuna-13B~\citep{vicuna2023}, in VQA tasks.
%
%
\item[$\bullet$]
Model designs that handle high-resolution image input~(e.g, dynamic high-resolution~\citep{ye2023ureader, liu2024llavanext, dong2024internlm-XComposer2, chen2024internvl1-5}) significantly boost performance on OCR-related tasks (e.g., OCRBench~\citep{liu2024OCRBench}), but sometimes show reduced accuracy on reasoning-related tasks (e.g., MMMU~\citep{yue2023mmmu}) compared to their low-resolution counterparts.
\item[$\bullet$]
Although open-access multimodal LLMs achieve impressive benchmark results on vision-language tasks, we observe a significant degradation in text-only performance~(see Table~\ref{tab:text-only}), unlike leading proprietary models (e.g., GPT-4o). 
The only work that provides substantial technical details addressing this issue is Llama 3-V~\citep{llama3-1}, which freezes the LLM parameters and trains only the cross-attention layers. However, these models have not yet been made publicly available.
\end{enumerate}

To address these limitations, we introduce NVLM-1.0, a family of frontier multimodal LLMs (see Figure~\ref{fig:Table0} for a comparison with leading models) featuring three distinct architectures: \emph{i)} NVLM-D, a \textbf{D}ecoder-only architecture, \emph{ii)} NVLM-X, a cross~(\textbf{X})-attention-based architecture, and \emph{iii)} NVLM-H, a novel \textbf{H}ybrid architecture.
Trained on the same curated data blend, all three architectures achieve state-of-the-art performance, rivaling leading proprietary and open-access models, while offering practitioners flexible and feature-rich model options. Specifically, \textbf{we make the following contributions}: 
\vspace{-0.4em}
\begin{enumerate}[leftmargin=1.7em]
    \item
    \textbf{Model architecture:}
     We compare the pros and cons of the decoder-only and the cross-attention-based models using the same LLM backbones, vision encoder, and well-curated training data. Our findings show that the cross-attention-based NVLM-X offers superior computational efficiency when handling high-resolution images, whereas the decoder-only NVLM-D provides unified multimodal reasoning and achieves higher accuracy in OCR-related tasks. Building on these insights, we propose NVLM-H, a novel hybrid architecture that excels in multimodal reasoning while also delivering improved computational efficiency for high-resolution images.
     \item 
    \textbf{High-resolution:}
    To achieve strong accuracy on both OCR-related tasks (e.g., OCRBench~\citep{liu2024OCRBench}) and multimodal reasoning tasks (e.g., MMMU~\citep{yue2023mmmu}), we propose a tile-tagging design for the dynamic tiling of high-resolution image inputs. Through comprehensive ablation studies, we find that adding a text-based 1-D tile tag before the image tokens of the corresponding tile in the decoder achieves the best accuracy.
    \item 
    \textbf{Training data:}
    We meticulously collect and provide detailed information on our multimodal pretraining and supervised fine-tuning (SFT) datasets, which will support and benefit future research.
    In the dataset selection and filtering process, we find that \emph{the data quality and task diversity are more important than the scale, even during the pretraining stage}. Furthermore, previous studies have shown that abundant and diverse multimodal pretraining data is crucial for the success of cross-attention-based models, such as Flamingo~\citep{alayrac2022flamingo}. In this work, we found that such pretraining data can also significantly improve the performance of decoder-only models, like LLaVA~\citep{liu2023improvedllava}, even with a simplified design that involves training only an MLP projection layer during pretraining.
    For the curation of SFT data, we collected a much larger set of task-oriented datasets compared to previous studies~\citep{chen2024internvl1-5}.
    \item 
    \textbf{Production-grade multimodality:}
     We develop \emph{production-grade multimodality} for NVLM models, enabling them to excel in both vision-language tasks (e.g., multimodal reasoning, OCR, natural image understanding) and text-only tasks (e.g., multidisciplinary knowledge reasoning, coding, and math). 
    To maintain text-only performance during multimodal training, we investigate two approaches:
    \emph{i)} For the cross-attention-based NVLM-X, we find that freezing the LLM’s parameters and training only the cross-attention layers~\citep{alayrac2022flamingo} during both the pretraining and SFT stages works reasonably well, with a moderate performance trade-off on vision-language tasks. 
    \emph{ii)}  We curate a high-quality text-only dataset and integrate it into the multimodal SFT stage, effectively preserving text-only performance with no degradation, and even achieving noticeable improvements on text-only math and coding benchmarks after multimodal training across all NVLM models.
    We attribute this to the superb quality of text-only data and the significant amount of multimodal math data~(e.g., geometry) incorporated into multimdoal SFT blend, which improves NVLM's reasoning capabilities, regardless of modality.
\end{enumerate}

We organize the rest of this paper as follows.
In \S~\ref{sec:qualitative_study}, we present a qualitative study of our model's capabilities, showcasing generated samples.
In \S~\ref{sec:preliminaries}, we introduce the preliminaries of multimodal LLMs and discuss related work.
In \S~\ref{sec:NVLM}, we present the NVLM-1.0 model family, followed by details on the training data in \S~\ref{sec:data}.
We introduce the evaluation benchmarks and report results in \S~\ref{sec:results}. 
We conclude the paper in \S~\ref{sec:conclusion}.

\section{Qualitative Study}
\label{sec:qualitative_study}
We conduct a qualitative analysis of NVLM-1.0 with diverse images and instructions.
As illustrated in Figure~\ref{fig:qualitative_example}, 
NVLM-1.0 can handle diverse types of images including memes in Figure~\ref{fig:qualitative_example}~(a), object-centric images in Figure~\ref{fig:qualitative_example} (b), real-world scene images in Figure~\ref{fig:qualitative_example} (c), hand-written pseudo code in Figure~\ref{fig:qualitative_example} (d), table in Figure~\ref{fig:qualitative_example} (e), and charts in Figure~\ref{fig:qualitative_example} (f).

Our NVLM-D$_{ 1.0}$72B demonstrates versatile capabilities in various multimodal tasks by jointly utilizing OCR, reasoning, localization, common sense, world knowledge, and coding ability.
For instance, our model can understand the humor behind the ``abstract vs. paper'' meme in Figure~\ref{fig:qualitative_example}~(a) by performing OCR to recognize the text labels for each image and using reasoning to grasp why 
juxtaposing ``the abstract'' — labeled with a fierce-looking lynx — and ``the paper'' — labeled with a domestic cat — is humorous.
NVLM accurately performs localization to effectively answer location-sensitive questions, such as ``What is the difference between the left, middle, and right objects in the image?'' in Figure~\ref{fig:qualitative_example}~(b). 
NVLM is capable of performing mathematical reasoning and coding based on visual information, such as tables and handwritten pseudocode, as illustrated in Figure~\ref{fig:qualitative_example}~(d) and~(e).
For more examples, refer to Appendix~\ref{appendix:more_examples} or our project site:~\url{https://nvlm-project.github.io/}.

\section{Preliminaries}
\label{sec:preliminaries}

Vision language models~\citep{radford2021learning, bao2021beit, wang2021simvlm, alayrac2022flamingo, wang2022image, zhu2023minigpt, chen2022pali, wang2022git, zhang2021vinvl} build the connection between the visual world and open text domain.
Among these works, the multimodal LLMs augmented from pretrained large language models (LLMs)~\citep{alayrac2022flamingo, awadalla2023openflamingo, yang2023re, li2022blip, li2023blip, dai2023instructblip, liu2023llava, liu2023improvedllava, liu2024llavanext, li2024llava-onevision, chen2024internvl, chen2024internvl1-5, 2024internvl2, lin2024vila, wang2023cogvlm, bai2023qwen-vl, chen2023minigptv2, tong2024cambrian, yao2024minicpm} have become visual assistants and universal task solvers for various vision-language tasks, including image / video captioning~\citep{lin2014microsoft, xu2016msr_vtt}, visual understanding and reasoning~\citep{yue2023mmmu}, chart and diagram-related QA~\citep{masry2022chartqa}, math reasoning in visual context~\citep{lu2024mathvista},  and optical character recognition~(OCR)~\citep{liu2024OCRBench}.

\subsection{Essential Building Blocks}
Multimodal LLM typically consists of two indispensable components: large language model~(LLM) and vision encoder.
\paragraph{Large Languge Model}
A multimodal LLM typically builds upon a text-only LLM for initialization. While there are exceptions where multimodal LLMs are pretrained from scratch using multimodal data~\citep{gemini1-0, adept2024-fuyu-heavy}, these approaches, though conceptually compelling, lack clear evidence of superior performance in vision-language tasks compared to multimodal LLMs built on a text-only LLM.

Instruction-tuned LLMs~\citep{wei2021finetuned, chung2024scaling, ouyang2022training} serve as universal task solvers in the text domain, as they can follow user-provided instructions to address a variety of tasks.
As a result, it is common to build multimodal LLMs on instruction-tuned LLMs rather than base LLMs in previous studies,~\citep{li2024llava-onevision, chen2024internvl1-5, lin2024vila, tong2024cambrian, llama3-1}, as the instruction-following capability is essential for solving a wide range of vision-language tasks.
Various instruction-tuned LLMs have been used to build multimodal LLMs in different study, including Vicuna-1.5~\citep{vicuna2023}, LLaMA-2-Chat~\citep{touvron2023llama2}, Mistral 7B~\citep{jiang2023mistral}, Yi-34B~\citep{young2024yi}, Llama3-Instruct~\citep{llama3-1}, and Qwen2-Instruct~\citep{qwen2}.
In this work, we use Qwen2-72B-Instruct~\citep{qwen2} as the default text-only LLM backbone.
We also employ Nous-Hermes-2-Yi-34B~\citep{Nous-Hermes-2-Yi-34B} for ablation study and faster experimentation. 

\paragraph{Vision Encoder.}
Multimodal LLMs~\citep[e.g.,][]{alayrac2022flamingo, li2024llava-onevision, li2023blip, chen2024internvl} typically leverage pretrained vision encoders (e.g., CLIP~\citep{radford2021learning}) to extract visual features from input images or video frames, with only a very few exceptions~\citep{adept2024-fuyu-heavy}. 
These vision encoders~\citep{radford2021learning, ilharco2021openclip, chen2024internvl, zhai2023SigLIP, dehghani2023scaling} are often trained on large-scale, diverse, and noisy text-image pairs sourced from the web~\citep{schuhmann2022laion, kakaobrain2022coyo-700m, gadre2024datacomp}.
This allows for large-scale training and enhances the generalization needed to effectively process visual input in unseen domains.
The other types of datasets, such as those used for optical character recognition (OCR)~\citep{chen2024internvl} and image segmentation~\citep{kirillov2023segment}, are also incorporated to enhance the specific capabilities of vision encoders.
In this study, we use InternViT-6B~\citep{chen2024internvl} as the default vision encoder due to its strength.
We keep this vision encoder frozen at all stages of training, as this simplifies the training process while still delivering strong results.

\subsection{Architectural Designs}
\label{subsec:prelim_arch}
There are various architectural designs for constructing multimodal LLMs (MLLMs) using existing LLMs and vision encoders~\citep{alayrac2022flamingo, li2022blip, liu2023llava, wang2023cogvlm, bai2023qwen-vl}. We discuss the two most common architectures.
\paragraph{Decoder-only MLLMs.}
Decoder-only architectures are popular mainly for their simplicity and unified handling of all modalities by aligning other modality tokens into the text token embedding space.
It also facilitates the extension to generating other modalities~\citep{gemini1-0, gpt4o}.
The notable examples of decoder-only multimodal LLMs include LLaVA~\citep{liu2023llava, liu2023improvedllava, liu2024llavanext, li2024llava-onevision}, InternVL~\citep{chen2024internvl, chen2024internvl1-5, 2024internvl2}, and Cambrian-1~\citep{tong2024cambrian}.
In these models, image tokens from the vision encoder are projected into the text-embedding space via a projector module, e.g., position-wise multi-layer perceptron (MLP), and then directly fed into the decoder-only LLM, just like the text tokens.
Some variants, such as Qwen-VL~\citep{bai2023qwen-vl}, utilize more advanced modules, e.g., \emph{Perceiver}~\citep{jaegle2021perceiver}, to down-sample the image tokens before they are fed into the LLM.

Training decoder-only multimodal LLMs typically involves two stages: \emph{pretraining} and \emph{supervised fine-tuning} (SFT). At the start of pretraining, the randomly initialized MLP or projector module needs to be trained while keeping the LLM frozen to avoid disrupting the LLM's weights~\citep{liu2023llava, liu2024llavanext}.  
Related work has also shown cases where both the projector and vision encoder are jointly trained during the pretraining stage~\citep{chen2024internvl1-5, bai2023qwen-vl}.
Due to the limited capacity of the MLP or projector module, the LLM need to be unfrozen during multimodal supervised fine-tuning~(SFT) to achieve good performance on vision-language tasks~\citep{lin2024vila}. 
The vision encoder is typically kept frozen during the SFT stage.
There are some exceptions, though, where the entire multimodal LLM is trained end-to-end~\citep{li2024llava-onevision}, usually with smaller vision encoder~\citep{zhai2023SigLIP}.

\paragraph{Cross-attention-based MLLMs.}
Cross-attention-based architectures are similar to encoder-decoder transformer models for machine translation~\citep{vaswani2017attention}, where the text decoder processes flattened image tokens via cross-attention layers, treating them as if they were a foreign language.
One of the early successful cross($X$)-attention architectures is Flamingo~\citep{alayrac2022flamingo}, which is built on frozen pretrained LLMs~\citep{hoffmann2022training} and often serves as the starting point for many studies on this type of model. 
The Flamingo model has two sets of trainable modules: \emph{i)} a \emph{perceiver resampler}~\citep{jaegle2021perceiver} positioned after the frozen vision encoder~\citep{radford2021learning}, which is designed to down-sample the vision encoder output to a specified size of representations, and \emph{ii)} the \emph{gated x-attention layers} interleaved with frozen LLM layers, which read output representations from the perceiver resampler.
In contrast, our NVLM-1.0-X and the concurrent Llama 3-V~\citep{llama3-1} models utilize only gated cross-attention layers to process image tokens and do not include the Perceiver module.

The Flamingo model was trained in two stages: 1) pretraining with a large (and possibly noisy) set of image-text pairs or interleaved image-text data, and  2) supervised fine-tuning~(SFT) with high-quality data.  It always freezes self-attention layers in LLM decoder and only trains cross-attention layers and perceiver during both pretraining and supervised fine-tuning~(SFT) to maintain text-only performance. 
At inference time, the gate of the $X$-attention layers can be turned ON for multimodal tasks and OFF for text-only tasks.
Thanks to the frozen LLM and gated $X$-attention designs, the text-only performance is guaranteed not to degrade after multimodal training.
The follow-up work includes IDEFICS~\citep{laurencon2023IDEFICS} and OpenFlamingo~\citep{awadalla2023openflamingo}, which are open-source reproductions of Flamingo.

In contrast to decoder-only models, cross-attention-based  MLLMs are generally considered more complex to implement. This complexity arises from the introduction of additional modules, the need for proper cross-attention masking in interleaved image-text settings, and the significantly heavier pretraining data requirements~\citep{alayrac2022flamingo, laurencon2023IDEFICS, zhu2023mmc4}.
However, a notable advantage of the $X$-attention-based architecture is its computational efficiency, as it does not require unrolling all image tokens in the LLM decoder, which typically results in long sequences during both training and inference, especially for high-resolution images.
See \S\ref{subsec:nvlm1.0x} for further study.

\subsection{High-Resolution Inputs}
\label{subsec:dynamic_high_res}
Properly handling high-resolution images is crucial for achieving state-of-the-art results in many OCR-related tasks.
However, vision encoders are typically trained with static resolution of  $224^2$ or $336^2$ pixels for efficiency~\citep{radford2021learning, openclip_repo}, when the image patch size per token is usually $14^2$ or $16^2$. For example, feeding a $224^2$ image to ViT-L/14~(patch size $14^2$) results in $(\frac{224}{14})^2=256$ tokens.
There are specialized vision encoders that can directly handle static high-resolution images. For instance, the SAM encoder~\citep{kirillov2023segment}, designed for image segmentation, can process images of $1024^2$ pixels with a ViT-L/16 backbone ($16^2$ pixels per patch), producing a 4096-token output. This can be costly, especially when training datasets and downstream tasks contain a mix of low-resolution and high-resolution images.

The dynamic high-resolution mechanism~\citep{ye2023ureader, liu2024llavanext, dong2024internlm-XComposer2, chen2024internvl1-5} has been proposed to address the waste of compute in such scenarios.
For example, given a ViT-L/14 vision encoder trained on low-resolution images (e.g., $224^2$), a high-resolution image (e.g., $896\times672$) is divided into tiles based on the aspect ratio and resolution of the input image ($\frac{896}{224}\times\frac{672}{224}=12$ tiles in this case). Each tile is independently fed into the ViT-L/14, producing 256 tokens per tile and 3072 tokens in total.
Meanwhile, it only produces 512 tokens for an input image with $448\times224$ resolution.
This dynamic approach is particularly well-suited for multimodal LLMs, which need to handle different types of tasks with varying image resolutions.

\begin{figure}
\centerline{\includegraphics[width=\linewidth]{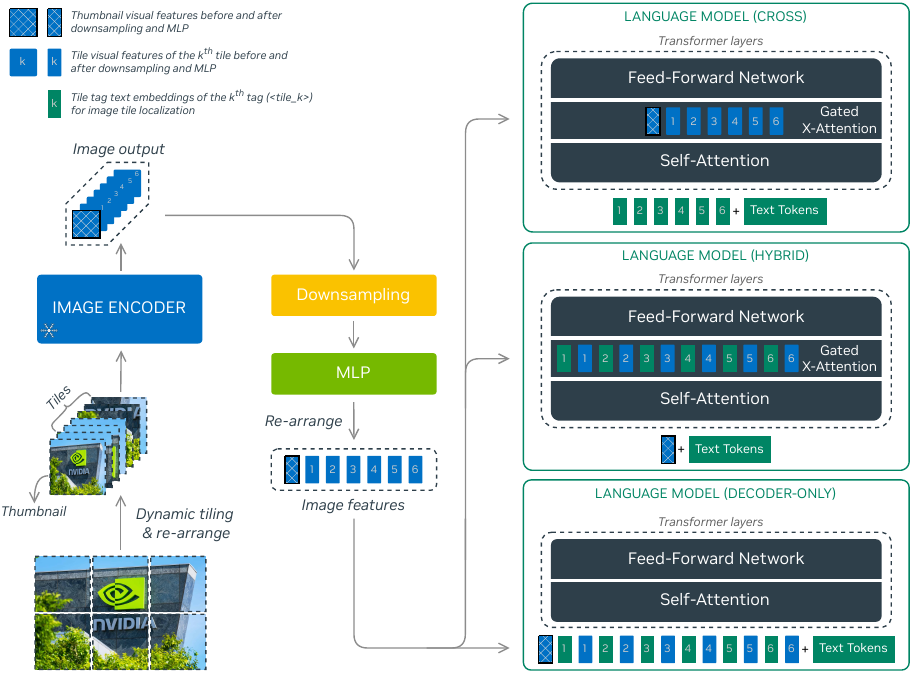}}
\vspace{.3cm}
\caption{\footnotesize
NVLM-1.0 offers three architectural options: the cross-attention-based NVLM-X (top), the hybrid NVLM-H (middle), and the decoder-only NVLM-D (bottom).
The dynamic high-resolution vision pathway is shared by all three models.
However, different architectures process the image features from thumbnails and regular local tiles in distinct ways.
}
\label{fig:nvlm-arch}
\end{figure}

\section{NVLM: Models and Training Methods}
\label{sec:NVLM}

In this section, we introduce NVLM-1.0, a family of frontier-class multimodal LLMs featuring three architectures: 
\emph{i)} \textbf{D}ecoder-only NVLM-D, 
\emph{ii)} Cross~(\textbf{X})-attention based NVLM-X, 
and \emph{iii)} NVLM-H with \textbf{H}ybrid architecture. 
Figure~\ref{fig:nvlm-arch} illustrates these architectures.
We will begin by detailing the vision pathway shared by all NVLM models.

\subsection{Shared Vision Pathway}
Several studies have compared various vision encoders in multimodal LLMs, suggesting that unfreezing and combining multiple smaller vision encoders offer advantages~\citep{tong2024cambrian}. In this work, we employ a single, large, and powerful vision encoder, InternViT-6B-448px-V1-5~\citep{InternViT-6B-448px-V1-5, chen2024internvl1-5}, as the default for all three architectures, keeping it frozen throughout all training stages. It processes images at a fixed resolution of $448^2$, generating 1,024 output tokens.

\begin{figure}
\centerline{\includegraphics[width=\linewidth]{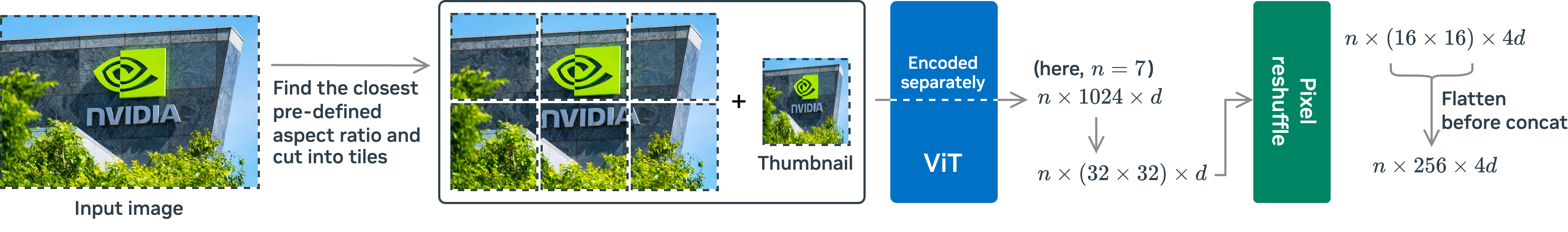}}
\caption{\footnotesize
Dynamic tiling of high-resolution input images.
Each tile is encoded separately by InternViT-6B, producing 1,024 tokens, which are downsampled to 256 tokens using a pixel shuffle operation.
}
\label{fig:dynamic-tile-details}
\end{figure}

We use the similar dynamic high-resolution~(DHR) approach outlined in \citet{chen2024internvl1-5}. See the left part of Figure~\ref{fig:nvlm-arch} for an illustration. 
We allow a maximum of 6 tiles at training. Thus, the predefined aspect ratios are: \{1:1, 1:2, 1:3, 1:4, 1:5, 1:6, 2:1, 2:2, 2:3, 3:1, 3:2, 4:1, 5:1, 6:1\}, encompassing all possible combinations of aspect ratios formed by 1 to 6 tiles.
For each input image, we dynamically match it to a predefined aspect ratio and divide it into 1 to 6 tiles, each corresponding to 448×448 pixels, based on the image's resolution.
We include a thumbnail tile, which is a scaled-down version of the entire image to capture the global context. 
Each tile is then fed into InternViT-6B-448px-V1-5~\citep{InternViT-6B-448px-V1-5}, generating 1,024 tokens.
We apply a \emph{downsampling} operation to reduce the 1,024 image tokens to 256, reducing the processing overhead for the LLM. This operation groups four neighboring image tokens into one by concatenating them along the channel dimension, a.k.a. pixel shuffle~\citep{chen2024internvl1-5}.
See Figure~\ref{fig:dynamic-tile-details} for a detailed illustration of this process.

This dynamic high-resolution~(DHR) design significantly improves performance on OCR-related tasks~\citep{chen2024internvl1-5, dong2024internlm-XComposer2}, but sometimes results in degraded results on reasoning-related tasks~\citep{yue2023mmmu} when all image tokens from the tiles are simply concatenated and fed directly into the LLM. 
We will address this issue across the three architectures, respectively.

\subsection{NVLM-D: Decoder-only Model}
\label{subsec:nvlm1.0d}
Similar to previous decoder-only multimodal LLMs~\citep{liu2023llava, chen2024internvl1-5}, NVLM-D model connects the pretrained vision encoder to the LLM using a 2-layer MLP as the projector or modality-alignment module.

Training NVLM-D involves two stages: pretraining and supervised fine-tuning (SFT). The MLP is randomly initialized and needs to undergo pretraining first, with both the vision encoder and LLM backbone kept frozen.
In our early exploration, we found that joint pretraining of the MLP projector and vision encoder is beneficial when the vision encoder is relatively weak (e.g., ViT-L/14~\citep{openclip_repo}) and the pretraining datasets are sufficiently diverse. However, after upgrading to the more powerful InternViT-6B-448px-V1-5~\citep{InternViT-6B-448px-V1-5}, the performance gains became marginal. Consequently, we opt to keep the vision encoder frozen during pretraining for the sake of simplicity.
During the SFT stage, both the MLP projector and LLM are trained to learn new vision-language tasks with novel instructions, while the vision encoder remains frozen. 
However, a less frequently discussed point in decoder-only MLLM literature is that leaving the LLM unfrozen during multimodal SFT training often results in significant degradation in text-only performance.
Our NVLM-D model effectively maintains text-only performance by incorporating a high-quality text-only SFT dataset.
The model configuration and training details for NVLM-D  models are in \S~\ref{subsec:model_config_training}.

\paragraph{Tile Tag for Dynamic High-Resolution.}
As illustrated in Figure~\ref{fig:nvlm-arch}, the LLM backbone needs to process the flattened image tokens from all dynamic high-resolution tiles, including an additional thumbnail tile. Directly concatenating flattened tokens without delimiters could confuse the LLM, as LLM lacks prior knowledge of the dynamic tiling process. 
To address this, we insert a text-based tile tag in the input sequence to signal the start of a tile and the position of this tile within the whole tiling structure.
After the tile tag, we append the flattened 256 image tokens of the tile.
Note that our design differs from previous work~\citep{dong2024internlm-XComposer2}, which globally flattens the image tokens from different tiles and inserts a newline symbol at the end of each row of tokens.
We observe improved results with our approach, particularly as we scale up the model size and training data.

\begin{table}[t]
\centering
\caption{\footnotesize
Ablation study of tile tag formats for dynamic high-resolution (DHR) using the decoder-only \textbf{NVLM-D} with Yi-34B as the backbone LLM.
All models are trained for 20K iterations with batch size 128 without checkpoint selection to ensure a straightforward comparison.
}
\vspace{.1cm}
\centering
\begin{adjustbox}{width={0.95\textwidth}}
\begin{tabular}{lccccccc}
\toprule
Tile tag format & MMMU (val)       & MathVista   & AI2D~(test)      & ChartQA        & DocVQA             & TextVQA         & OCRBench    
\\ \midrule
Low-resolution~($448^2$)          & 50.9          & 46.1           & 67.0        & 64.8          & 52.9         & 78.2         &  622          \\ 
DHR + No tag          & 50.0          & 51.7      & 79.9      & 76.1          & 80.2          & 78.4          & 728           \\ 
DHR + 2-D grid tag    & 51.1          & 52.8   & 81.7                 &  {81.1} & 86.7          & 79.4         & 787           \\
DHR + 2-D bbox tag    & 50.3          & 50.6     &  81.2       & 80.8          & 86.7          & 79.7           & 791         \\ 
\textbf{DHR + 1-D tag}         & {52.0} & {53.8} & {82.1} & {81.1} & {87.4} & {79.9}  & {806} \\
\bottomrule
\end{tabular}
\end{adjustbox}
\label{tab:nvlm-d-ablation-on-tile-tag}
\end{table}

We introduce three different tile tags, and perform an ablation study on NVLM-D with Yi-34B~\citep{Nous-Hermes-2-Yi-34B} as the LLM backbone using the following variants of tile tags:

\begin{enumerate}[leftmargin=0.7cm]
  \item[\textit{a)}] No tag: Simple concatenation without tile tag, which is the design of InternVL-1.5~\citep{chen2024internvl1-5}.
  \item[\textit{b)}] 1-D flattened tile tag: {\footnotesize <$\mathtt{tile\_1}$>,~ <$\mathtt{tile\_2}$>,~ $\cdots$,~ <$\mathtt{tile\_6}$>,~ <$\mathtt{tile\_global}$>.
  }
  \item[\textit{c)}]  2-D grid tag: {\footnotesize <$\mathtt{tile\_x0\_y0}$>,~ <$\mathtt{tile\_x1\_y0}$>,~ $\cdots$,~ <$\mathtt{tile\_xW\_yH}$>,~ <$\mathtt{tile\_global}$>, where the \{$\mathtt{i : j}$\} of <$\mathtt{tile\_xi\_yj}$> can be in \{1:1, 1:2, 1:3, 1:4, 1:5, 1:6, 2:1, 2:2, 2:3, 3:1, 3:2, 4:1, 5:1, 6:1\}.
  }
  \item[\textit{d)}] 2-D bounding-box tag: {\footnotesize <$\mathtt{box}$> ($x_0$, $y_0$), ($x_1$, $y_1$) </$\mathtt{box}$>, $\cdots$, <$\mathtt{box}$> ($x_W$, $y_H$), ($x_{W+1}$, $y_{H+1}$) </$\mathtt{box}$>, where the ($x_i$, $y_j$), ($x_{i+1}$, $y_{j+1}$) are the (left, top), (right, bottom) coordinates of that particular title within the whole high-resolution image.
  }
\end{enumerate}

From Table~\ref{tab:nvlm-d-ablation-on-tile-tag}, we can observe that:
{1)} The vanilla dynamic high-resolution method~(DHR + No tag) significantly improves performance across all benchmarks, except for MMMU (50.0 vs. 50.9), compared to its low-resolution counterpart.
It is worth mentioning that previous DHR methods~\citep{dong2024internlm-XComposer2, chen2024internvl1-5} also exhibit lower MMMU accuracy compared to their low-resolution counterparts. 
{2)} Inserting all types of tile tags into the LLM decoder significantly outperforms simple concatenation with no tags.
 In particular, we find that the introduction of tile tag greatly improves the performance on OCR-related tasks, including ChartQA~\citep{masry2022chartqa}, DocVQA~\citep{mathew2021docvqa} and OCRBench~\citep{liu2024OCRBench}.
{3)} 1-D tile tag <$\mathtt{tile\_k}$> performs generally better than other tags. We hypothesize that although the 1-D tile tag does not tell 2-D information~(e.g., 2$\times$3 vs. 3$\times$2), it offers better generalization at test time. 
Importantly, this tile tag design for dynamic high-resolution also offers moderate improvement on math and multidisciplinary reasoning tasks, including MathVista~\citep{lu2024mathvista} and MMMU~\citep{yue2023mmmu}.

\subsection{NVLM-X: X-attention Model}
\label{subsec:nvlm1.0x}
NVLM-X employs gated cross-attention to process image tokens and differs from the Flamingo model~\citep{alayrac2022flamingo} in two key ways:
\begin{enumerate}[leftmargin=1em]
\item[\textbullet]
During our initial exploration, we found that while the \emph{perceiver resampler} is beneficial for natural image captioning, it negatively impacts dense OCR tasks, such as transcribing text from scanned documents (see Appendix~\ref{appendix:perceiver_in_flamingo} for further details). The primary reason is that the cross-attention to latent array in the Perceiver~\citep{jaegle2021perceiver} mixes the input image tokens, potentially disrupting the spatial relationships between image patches, which are crucial for document OCR.
Based on this observation, our NVLM-X architecture does not use a perceiver resampler; instead, it relies solely on cross-attention to read image tokens directly from the vision encoder.
\item[\textbullet]
Freezing the LLM during the multimodal SFT stage compromises performance on vision-language tasks, as the multimodal LLM needs to quickly adapt to new tasks and novel instructions that were not encountered during text-only instruction tuning.
We illustrate this observation in Table~\ref{tab:frozen_unfrozen_LLM_study} in \S~\ref{sec:results}.
Thus, we unfreeze the LLM backbone of NVLM-X during multimodal SFT  and blend in a high-quality text-only SFT dataset to maintain strong text-only performance.
Note that this also differs from Llama 3-V~\citep{llama3-1}, which freezes the LLM during multimodal training.
\end{enumerate}

The model configuration and training details for NVLM-X models can be found in \S~\ref{subsec:model_config_training}.

\paragraph{Tile Tag for Dynamic High-Resolution.}
NVLM-X uses the same dynamic high-resolution approach as NVLM-D to obtain image tokens from a global thumbnail tile and regular tiles. 
As illustrated in Figure~\ref{fig:nvlm-arch}, NVLM-X employs gated $X$-attention to process the flattened image tokens for each tile, rather than feeding them directly into the LLM decoder.
Similar to the design used in NVLM-D,  we insert a sequence of text-based tile tags <$\mathtt{tile\_1}$> $\cdots$  <$\mathtt{tile\_k}$>  in the LLM decoder, while allowing each tag <$\mathtt{tile\_k}$> to only attend to its corresponding image tokens by properly configuring the $X$-attention mask. 
This approach ensures that the LLM is better informed about the tiling structure without needing to infer it from the content of thumbnail tile and regular tiles.

In Table~\ref{tab:nvlm-x-ablation-on-tile-tag}, we present an ablation study of NVLM-X with Yi-34B LLM backbone using low-resolution $448^2$ input, dynamic high-resolution~(DHR) without tile tags and with 1-D <$\mathtt{tile\_k}$>~ tags.
We find that:
{1)}~The vanilla dynamic high-resolution approach (DHR + No tag) significantly outperforms its low-resolution counterpart across all benchmarks, except MMMU (53.0 vs. 53.2).
{2)}~Adding tile tags further improves performance across all benchmarks, including multimodal reasoning (MMMU: 54.1 vs. 53.0, MathVista: 59.6 vs. 57.6) and OCR-related tasks.

\begin{table}[t]
\centering
\caption{\footnotesize
Ablation study of using tile tag ~<$\mathtt{tile\_k}$>~ for dynamic high-resolution~(DHR) using the cross-attention-based \textbf{NVLM-X} with Yi-34B as the backbone LLM.
All models are trained for 10K iterations with batch size 512 without checkpoint selection to ensure a straightforward comparison.
}
\vspace{.1cm}
\centering
\begin{adjustbox}{width={\textwidth}}
\begin{tabular}{lccccccc}
\toprule
Tile-tag format & MMMU (val)       & MathVista     & AI2D~(test)          & ChartQA        & DocVQA     & TextVQA        & OCRBench         \\ \midrule
Low-resolution~($448^2$) & 53.2 & 57.4 & 69.8 & 65.6 & 62.3  & 63.2 & 612        \\ 
DHR + No tag &  53.0 & 57.6 & 78.5 & 73.2 & 75.4  & 75.2 & 682    \\ 
\textbf{DHR + 1-D tag} &  54.1 & 59.6  & 80.7 & 79.2 & 80.9 & 78.6 & 744    \\ \bottomrule
\end{tabular}
\end{adjustbox}
\label{tab:nvlm-x-ablation-on-tile-tag}
\end{table}

\paragraph{Decoder-only vs. X-attention.}
The pros and cons of  cross-attention-based NVLM-X and decoder-only NVLM-D can be summarized in the following.
\emph{i) Parameter efficiency:} NVLM-D has fewer parameters than NVLM-X, as the latter has the newly introduced gated cross-attention layers.
The number of additional parameters becomes significant as the model scales up. For instance, Llama 3-V 405B added 100B parameters to the text-only Llama-3.1-405B-Instruct~\citep{llama3-1}.
\emph{ii)} \emph{Training efficiency:}
NVLM-X enables more efficient processing of high-resolution images by eliminating the need to unroll all image tokens on the LLM decoder side. See Table~\ref{tab:traning_speed_nvlm} for a comparison of training \emph{throughput} between 34B NVLM-D and NVLM-X models. Note that the decoder-only NVLM-D requires much longer sequence lengths, as all image tokens are concatenated and fed into the LLM decoder, leading to higher GPU memory consumption and lower training throughput.
\emph{iii) Multimodal reasoning:}  NVLM-D performs unified processing of all tokens from different modalities, enabling joint multimodal reasoning at the LLM decoder. However, the long sequence of tokens for high-resolution images~(e.g., 256$\times$7 = 1792 tokens) may still make reasoning challenging, even with the assistance of tile tags.

\begin{table}[!t]
\centering
\caption{\footnotesize
Training throughput (samples per second) of NVLM-X, NVLM-D, and NVLM-H with Yi-34B as the backbone LLM.
We use 128 H100 GPUs during supervised fine-tuning with unfrozen LLMs.
All three models are implemented in Megatron-LM with tensor parallelism set to 8~\citep{shoeybi2019megatron}. Sequence lengths in the LLM decoder are set with 1,024 tokens for text, 256 image tokens for the thumbnail tile, and 256×6 for the 6 regular tiles.
}
\vspace{0.1cm}
\begin{adjustbox}{width={1.0\textwidth}}
\begin{tabular}{lcclccc}
\toprule
\multirow{2}{4em}{Models}  & \multirow{2}{5em}{Batch size}  & \# of H100   &  Sequence length  & \multirow{2}{4em}{\# of Tiles}  & Elapsed time (ms)  &  Throughput \\ 
& & GPUs & in LLM decoder  & &  per iteration & samples / sec  \\
\midrule
NVLM-X 34B        & 256   & 128     & 1,024   & 6+1  &  5,063 & 50.6 \\  
NVLM-D 34B  &  256 & 128 &  1,024 + 256$\times$7 = 2,816  & 6+1 & 8,885 & 28.8 \\
NVLM-H 34B  &  256 & 128  & 1,024 + 256 = 1,280 & 6+1 & 7,071  & 36.2 \\
\bottomrule
\end{tabular}
\end{adjustbox}
\label{tab:traning_speed_nvlm}
\end{table}

\subsection{NVLM-H: Hybrid Model}
\label{subsec:nvlm1.0h}

Drawing inspiration from the comparison of NVLM-X and NVLM-D, we propose NVLM-H, a novel hybrid architecture that combines the best of both approaches.
As illustrated in Figure~\ref{fig:nvlm-arch}, NVLM-H separates the processing of image tokens into two paths.
The thumbnail image tokens are fed into the LLM alongside text tokens and processed by self-attention layers, enabling joint multimodal reasoning.
Simultaneously, a dynamic number of regular tiles are processed through gated cross-attention, enabling the model to capture finer image details.
This approach enhances high-resolution capability compared to NVLM-X while significantly improving computational efficiency compared to NVLM-D.
Table~\ref{tab:traning_speed_nvlm} demonstrates that NVLM-H has higher throughput than NVLM-D in training.


\paragraph{Tile Tag for Dynamic High-Resolution.} 
NVLM-H utilizes the same 1-D flattened tile tag <$\mathtt{tile\_k}$> introduced in \S\ref{subsec:nvlm1.0d} for NVLM-D. The primary distinction lies in the processing location. As shown in Figure~\ref{fig:nvlm-arch}, text embeddings of   <$\mathtt{tile\_k}$>  are integrated into the gated cross-attention layers alongside visual embeddings. This approach is effective because the text and visual embeddings are well-aligned during pre-training, enabling the model to seamlessly interpret tile tags within the cross-attention mechanism. Consistent with the results in Table \ref{tab:nvlm-d-ablation-on-tile-tag} and Table \ref{tab:nvlm-x-ablation-on-tile-tag}, adding tile tags enhances NVLM-H's performance on OCR-related tasks compared to no tagging. 

\subsection{Model Configurations and Training Method}
\label{subsec:model_config_training}
We provide the model configurations and training details for all NVLM-1.0 models below.
\paragraph{Backbone LLMs and Vision Encoder.}
For the NVLM-D, NVLM-X, and NVLM-H 72B models, we use Qwen2-72B-Instruct~\citep{qwen2} as the backbone LLM. 
For computational reasons, we also use the smaller Nous-Hermes-2-Yi-34B~\citep{Nous-Hermes-2-Yi-34B} as the LLM backbone for faster ablation studies and experimentation.
After finalizing the optimized designs, we shifted our computational resources to improving the NVLM-1.0 72B models.
Across all NVLM models, InternViT-6B-448px-V1-5~\citep{InternViT-6B-448px-V1-5} serves as the vision encoder.

\paragraph{Modality-Alignment Module.}
We include the details of modality-alignment modules for three NVLM architechtures in the following:
\begin{itemize}[leftmargin=1.1em]
    \item 
    For NVLM-D models, the LLM and vision encoder are connected by a two-layer MLP to align the modalities, with hidden dimensions of $12800 \rightarrow 20480 \rightarrow 7168$ for 34B model, and $12800 \rightarrow 29568 \rightarrow 8192$ for 72B model. 
    Note that InternViT-6B has a hidden dimension of 3200, which increases to $3200\times4=12800$ after applying pixel shuffle. 
    Yi-34B~\citep{Nous-Hermes-2-Yi-34B} has hidden dimension $7168$, and Qwen2-72B has hideen dimension $8192$.
    \item 
    For NVLM-X models, the images features are first projected to LLMs' hidden dimension with a one-layer MLP,  $12800 \rightarrow 7168$ for 34B model, and $12800 \rightarrow 8192$ for 72B model. 
    We insert a gated $X$-attention layer every 6 and 8 LLM self-attention layers, respectively.  This results in a total of 10 $X$-attention layers for both models.
    \item 
    The NVLM-H 34B and 72B models utilize a two-layer MLP and $X$-attention layers as the modality-alignment module. The image tokens for both thumbnail and regular tiles are projected through the two-layer MLP, with hidden dimensions of $12800 \rightarrow 20480 \rightarrow 7168$ for the 34B model, and $12800 \rightarrow 29568 \rightarrow 8192$ for the 72B model. The projected thumbnail image tokens are then directly fed into the LLM decoder. The projected image tokens of regular tiles are cross-attended by the $X$-attention layers. As with NVLM-X, ten gated $X$-attention layers are inserted for both the 34B and 72B models.
\end{itemize}

\paragraph{Training Method.}
We employ a unified training method for all NVLM models.
The training process involves two stages:
\emph{i)} \textbf{Pretraining}: we freeze both the LLM backbone and vision encoder for all models. We only train the modality-alignment modules, i.e., projector MLP or $X$-attention layers, using our pretraining dataset detailed in Table \ref{tab:pretraining_data}. 
For pretraining hyperparameters, one can refer to Table~\ref{tab:hyperparams_pretraining} in  Appendix~\ref{appendix:training_configuration}.
We find a large batch size of 2048 improves the pretraining with frozen LLMs.
\emph{ii)} \textbf{Supervised fine-tuning}~(SFT): we keep the vision encoder frozen while training both the LLM and modality-alignment modules with our multimodal SFT datasets detailed in Table~\ref{tab:sft_data}, along with a text-only SFT dataset.
For hyperparameters of SFT,  one can refer to Table~\ref{tab:hyperparams_sft} in  Appendix~\ref{appendix:training_configuration}.

\section{Training Data}
\label{sec:data}

In this section, we provide details of the pretraining and supervised fine-tuning (SFT) datasets. These curated training datasets are used across all three architectures in the NVLM family.
All datasets are formatted based on the task type and the chat template provided in Appendix~\ref{appendix:chat_template}.

\subsection{Multimodal Pretraining Data}
We curate a diverse, high-quality multimodal pretraining dataset, all sourced from the open-source community.
We find that \emph{the quality of the dataset matters more than its scale, even at the pretraining stage}.
In early exploration, we experimented with much larger but noisier datasets~\citep[e.g.,][]{schuhmann2022laion, gadre2024datacomp}, commonly used for training CLIP-style vision encoders.
However, we found these unfiltered datasets to be less effective for training both decoder-only and $X$-attention-based multimodal LLMs, even with a frozen LLM during pretraining. 
The potential reason could be that noisy text-image data leads to inaccurate alignment between the two modalities.
We also experimented with interleaved text-image datasets, including MMC4~\citep{zhu2023mmc4} and OBELICS~\citep{laurenccon2024obelics}, and found that they had minimal impact on downstream vision-language tasks, even in few-shot settings, within the state-of-the-art NVLM framework.
We hypothesize more careful filtering and recaptioning are needed for such interleaved text-image datasets.

\begin{table}[!t]
\centering
\caption{\footnotesize
Datasets used by NVLM-1.0 at the pretraining stage.
}
\vspace{.1cm}
\label{tab:pretraining_data}
\begin{adjustbox}{width={1.0\textwidth}}
\begin{tabular}{ll}
\toprule
Task   &  Dataset \\ \midrule
\multirow{1}{4em}{Captioning}  & COCO~\citep{lin2014microsoft}, CC3M~\citep{sharma2018conceptual}, SBU~\citep{ordonez2011im2text}, LAION-115M~(sanitized)~\cite{schuhmann2021laion, li2022blip} 
\\  \midrule
VQA (natural image)  & VQAv2~\citep{goyal2017making},  Visual Genome~\citep{krishna2017visual} 
\\ \midrule
Chart &  DVQA~\citep{kafle2018dvqa}  
\\ \midrule
Document & Docmatix~\citep{Docmatix2024}  
\\  \midrule
\multirow{2}{6em}{OCR / Scene-Text}    & OCR-VQA~\citep{mishra2019ocr}, COCO-Text~\citep{veit2016cocotext},  TextOCR~\citep{singh2021textocr}, ReCTs~\citep{zhang2019icdar}, RRC-ArT~\citep{chng2019icdar2019}, RRC-LSVT~\citep{sun2019icdar}  \\
& RCTW~\citep{shi2017icdar2017}, synthdog-en~\citep{kim2022donut}, pdfa-eng-wds~\citep{pdfa-eng-wds}
\\ \midrule
Math &   CLEVR-Math~\citep{lindstrom2022clevr}
\\ \bottomrule
\end{tabular}
\end{adjustbox}
\end{table}

\begin{table}[t]
\centering
\vspace{-.3cm}
\caption{\footnotesize
An ablation study comparing the use of our pretraining data in Table~\ref{tab:pretraining_data} with the pretraining data from LLaVA-1.5~\citep{LLaVA-Pretrain} using decoder-only NVLM-D with Yi-34B as the backbone LLM. Both models are trained for 20K iterations with batch size 128 without checkpoint selection to ensure a straightforward comparison.
}
\vspace{.1cm}
\centering
\begin{adjustbox}{width={\textwidth}}
\begin{tabular}{lccccccc}
\toprule
Pretraining data & MMMU (val)       & MathVista     & AI2D~(test)          & ChartQA        & DocVQA     & TextVQA        & OCRBench         \\ \midrule
LLaVA-1.5 data~\citep{LLaVA-Pretrain}  &  51.8 & 48.9 & 80.5 & 80.3 & 85.2  & 78.9 &  760   \\ 
Our pretraining data &  {52.0} & {53.8} & {82.1} & {81.1} & {87.4} & {79.9}  & {806} \\ \bottomrule
\end{tabular}
\end{adjustbox}
\label{tab:ablation_on_pretraining_data}
\end{table}

We provide a list of the pretraining datasets in Table~\ref{tab:pretraining_data}. 
These datasets cover different tasks: 
1)~Captioning. In particular, we use a filtered and recaptioned version of LAION-115M from \citet{li2022blip}. We perform thorough data scanning and sanitization to ensure the dataset is free of any harmful or inappropriate content.
2)~Visual question answering~(VQA) on natural image.
3)~VQA on chart and scanned document.
4)~Math reasoning in a visual context.
5)~OCR and scene-text recognition.
In addition to large-scale captioning datasets, we find that incorporating large task-oriented datasets during the pretraining stage enhances cross-modal alignment and leads to better final results. We also experimented with blending relatively small task-oriented datasets used in SFT into pretraining. However, this approach caused overfitting on these datasets and impaired the model's reasoning ability when evaluated on zero-shot benchmarks such as MMMU and MathVista.

Previous work~\citep{alayrac2022flamingo} has shown that abundant and diverse pretraining data is crucial for the success of cross-attention-based models.
In contrast, decoder-only models, such as LLaVA~\citep{liu2023improvedllava, liu2024llavanext}, work well with smaller pretraining datasets~\citep{LLaVA-Pretrain}, which are simply filtered subsets of the captioning datasets including CC3M~\citep{sharma2018conceptual}, SBU~\citep{ordonez2011im2text}, and LAION-115M~\cite{schuhmann2021laion, li2022blip}.
In contrast, our findings demonstrate that the diverse pretraining data shown in Table~\ref{tab:pretraining_data}
 can still significantly enhance the performance of decoder-only multimodal LLMs, even in state-of-the-art settings with highly curated SFT datasets.
We conducted an ablation study comparing our pretraining data with LLaVA-1.5's pretraining data~\citep{LLaVA-Pretrain} for the NVLM-D with Yi-34B as LLM backbone, as shown in Table~\ref{tab:ablation_on_pretraining_data}. The pretrained models are then fine-tuned on the same high-quality SFT dataset in Table~\ref{tab:sft_data}.
One can see that our diverse pretraining data provide consistent improvements across all benchmarks, in particular a significant improvement in math reasoning and OCR-related tasks, as we add these types of data in pretraining.

\begin{table}[!t]
\centering
\caption{\footnotesize
Datasets used by NVLM-1.0 at supervised-fine-tuning~(SFT).
}
\vspace{.1cm}
\label{tab:sft_data}
\begin{adjustbox}{width={1.0\textwidth}}
\begin{tabular}{ll}
\toprule
\textbf{Task}   &  \textbf{Dataset} \\ \midrule
\multirow{1}{4em}{Captioning}  & COCO~\citep{lin2014microsoft}, TextCaps~\citep{sidorov2020textcaps}, ShareGPT-4o~\citep{ShareGPT-4o}
\\  \midrule
VQA (natural image)  & VQAv2~\citep{goyal2017making}, Visual Genome~\citep{krishna2017visual}, TallyQA~\citep{acharya2019tallyqa}, Visual7W~\citep{zhu2016visual7w}, 
Vizwiz~\citep{gurari2018vizwiz}
 \\ \midrule
General Knowledge  &  OK-VQA~\citep{marino2019ok}, A-OKVQA~\citep{schwenk2022okvqa}
\\  \midrule
Visual Reasoning &  GQA~\citep{hudson2019gqa}, Super-CLEVR~\citep{li2023super-clevr}, Raven~\citep{zhang2019raven}, 
VSR~\citep{liu2023visual}
\\ \midrule
\multirow{2}{7em}{Chart \& Diagram} &  DVQA~\citep{kafle2018dvqa}, PlotQA~\citep{methani2020plotqa}, MMC-Instruction~\citep{liu2023mmc}, ChartQA~\citep{masry2022chartqa}, InfographicVQA~\citep{mathew2022infographicvqa}
\\
& FigureQA~\citep{kahou2017figureqa},  IconQA~\citep{lu2021iconqa}, Chart2Text~\citep{obeid2020chart2text}, Diagram Image2Text~\citep{diagram-image2text}
\\ \midrule
Table &  WikiTableQuestions~\citep{pasupat2015compositional},
RobuT(WTQ, WikiSQL, SQA)~\citep{zhao2023robut}, HiTab~\citep{cheng2021hitab}
\\ \midrule
\multirow{2}{4em}{Document}  &  DocVQA~\citep{mathew2021docvqa}, Docmatix~\citep{Docmatix2024},  DUDE~\citep{van2023document},  VisualMRC~\citep{tanaka2021visualmrc}, TAT-DQA\citep{zhu2022towards} \\
&  UReader IE~\citep{ye2023ureader}, UReader KG~\citep{ye2023ureader}, UReader QA~\citep{ye2023ureader}, 
\\ \midrule
\multirow{5}{6em}{OCR / Screen / Scene-Text}  & OCR-VQA~\citep{mishra2019ocr},  
 TextVQA~\citep{singh2019towards}, ST-VQA~\citep{biten2019scene}, ScreenQA~\citep{hsiao2024screenqa},  SlideQA~\citep{SlideVQA2023},  PDF-VQA~\citep{ding2023pdfvqa}  \\
 & VQA-CD~\citep{mahamoud2022VQA-CD}, VQAonBD~\citep{VQAonBD2023}, POIE~\citep{kuang2023visual}, SROIE~\citep{huang2019icdar2019}, ORAND~\citep{diem2014icfhr}, 
 EST-VQA~\citep{wang2020general}
 \\ 
 & FUNSD~\citep{jaume2019funsd}, SQuAD(rendering)~\citep{rajpurkar2016squad}, WordArt~\citep{xie2022toward}, IAM~\citep{marti2002iam}, IIIT5K~\citep{IIIT5K},  HME100K~\citep{yuan2022syntax}
 \\
 & synthdog-en~\citep{kim2022donut}, Bentham QA~\citep{mathew2021asking}, HW-SQuAD~\citep{mathew2021asking},
 WebSight~\citep{laurenccon2024unlocking}, ChromeWriting~\citep{chromewriting}  \\
 & K12 Printing~\citep{li2024llava-onevision}, COCO-Text~\citep{veit2016cocotext}, TextOCR~\citep{singh2021textocr}, ReCTs~\citep{zhang2019icdar}, pdfa-eng-wds~\citep{pdfa-eng-wds}
 \\  \midrule
\multirow{3}{4em}{Math} &   CLEVR-Math~\citep{lindstrom2022clevr},  GeoQA+~\citep{cao2022augmented}, Geometry3K~\citep{lu2021inter},  TabMWP~\citep{lu2022dynamic}, GSM8K(rendering)~\citep{cobbe2021training}  \\
& MetaMathQA(rendering)~\citep{yu2023metamath}, MAVIS Data Engine~\citep{zhang2024mavis}, MAVIS Manual Collection~\citep{zhang2024mavis} \\
& Geo170K Align~\citep{gao2023g}, Geo170K QA~\citep{gao2023g}, GeoMVerse~\citep{kazemi2023geomverse}, GEOS~\citep{seo2015solving}, UniGeo~\citep{chen2022unigeo}
\\ \midrule
Science &  AI2D~\citep{kembhavi2016diagram},  ScienceQA~\citep{lu2022learn}, TQA~\citep{kembhavi2017you}, 
ArXivQA~\citep{li2024multimodal}, textbook data \\ \midrule
 Visual Instruction-Tuning &  LRV-Instruction~\citep{liu2023aligning}, LLaVA-158K~\citep{liu2023llava}, LLaVAR~\citep{zhang2023llavar}
 \\ \midrule
\multirow{3}{6em}{Text-only SFT}  & SlimOrca~\citep{SlimOrca}, ShareGPT~\citep{ShareGPT_Vicuna_unfiltered}, EvolInstruct~\citep{xu2024wizardlm}, GPTeacher~\citep{GPTeacher2023}, AlpacaGPT4~\citep{peng2023instruction}, \\ 
& UltraInteract~\citep{yuan2024advancing}, OrcaMathWordProblems~\citep{mitra2024orca}, MathInstruct~\citep{yue2023mammoth}, MetaMath~\citep{yu2023metamath}, \\ 
&  GlaiveCodeAssistant~\citep{GlaiveCodeAssistant2023}, Magicoder~\citep{wei2024magicoder}, WizardCoder~\cite{luo2023wizardcoder}. \\
 \bottomrule
\end{tabular}
\end{adjustbox}
\end{table}

\subsection{Multimodal SFT Data}
We collected a diverse, high-quality, task-oriented SFT dataset to enhance NVLM's capabilites on a wide range of visoin-language tasks.
A detailed list of SFT datasets is provided in Table~\ref{tab:sft_data}.
In addition to high-quality datasets with short captions, such as COCO~\citep{lin2014microsoft} and TextCaps~\citep{sidorov2020textcaps}, we also include ShareGPT-4o~\citep{ShareGPT-4o}, which provides detailed image descriptions.
Additionally, we have included several VQA datasets based on natural images~\citep{goyal2017making}, with a focus on object layout~\citep{krishna2017visual}, counting~\citep{acharya2019tallyqa}, object-level grounding~\citep{zhu2016visual7w}, mobile phone photo with varying quality~\citep{gurari2018vizwiz}, visual reasoning~\citep{hudson2019gqa, li2023super-clevr, zhang2019raven, liu2023visual}, and knowledge-based VQA~\citep{marino2019ok, schwenk2022okvqa}.
The ability to understand charts, diagrams, tables, document images is a critical real-world application of multimodal LLMs. To enhance this capability, we have incorporated a diverse set of datasets~(e.g., DVQA~\citep{kafle2018dvqa}, PlotQA~\citep{methani2020plotqa}, WikiTableQuestions~\citep{pasupat2015compositional}, DocVQA~\citep{mathew2021docvqa}). 
OCR is a fundamental capability of multimodal LLMs, as it is directly related to performance on tasks involving scene text~\citep{singh2019towards}, screenshots~\citep{hsiao2024screenqa}, charts~\citep{masry2022chartqa}, tables~\citep{cheng2021hitab}, document images~\citep{mathew2021docvqa}, and handwritten text.
As a result, we have incorporated a substantial amount of OCR-related datasets in our SFT blend.
Another important capability is mathematical reasoning within a visual context. To enhance this, we have incorporated many multimodal math reasoning datasets listed in Table~\ref{tab:sft_data}.
Interestingly, the abundant multimodal math data not only leads to significant improvements in vision-language tasks like MathVista~\cite{lu2024mathvista}, but also results in substantial gains on text-only math benchmarks, including GSM8K~\citep{cobbe2021gsm8k} and MATH~\citep{hendrycks2021math}.
Following previous leading open-source work~\citep[e.g.,][]{chen2024internvl1-5, li2024llava-onevision}, we incorporate the training splits of datasets including ChartQA~\citep{masry2022chartqa}, DocVQA~\citep{mathew2021docvqa}, VQAv2~\citep{goyal2017making}, TextVQA~\citep{singh2019towards} and AI2D~\citep{kembhavi2016diagram} into the SFT blend. Their test sets are used as evaluation benchmarks in Section~\ref{sec:results}, meaning they are not evaluated in a zero-shot setting.
\textbf{
We want to emphasize that we did not apply data augmentation to the training splits of these benchmark datasets. While such techniques certainly improve benchmark results, they contribute minimal value to the model's out-of-domain generalization.}
Note that it is unknown whether the proprietary multimodal LLMs are being evaluated on these benchmarks in a zero-shot or fine-tuning setting, as no information is provided regarding their training datasets. We hypothesize that it is a fine-tuning setting, based on observed accuracy gaps between the training and test sets for some proprietary models; however, this is not conclusive.

\subsection{Text-only SFT Data}
We curated a high-quality text-only SFT dataset and incorporated it into the multimodal fine-tuning stage, effectively preserving the LLM backbone's text-only performance and preventing catastrophic forgetting. Previous leading open-access multimodal LLMs~\citep{chen2024internvl1-5, li2024llava-onevision} also include text-only SFT datasets but still show significant performance degradation on text-only benchmarks (see Table~\ref{tab:text-only} for details). The key difference between our recipe and theirs lies in the quality of the data.

Our text-only SFT dataset is built on top of open-source SFT datasets. We collect SFT datasets from general categories, including ShareGPT~\citep{vicuna2023, ShareGPT_Vicuna_unfiltered}, SlimOrca~\citep{SlimOrca,mukherjee2023orca}, EvolInstruct~\citep{xu2024wizardlm}, GPTeacher~\citep{GPTeacher2023}, AlpacaGPT4~\citep{peng2023instruction}, and UltraInteract~\citep{yuan2024advancing}. Additionally, we collect datasets from math category, including OrcaMathWordProblems~\citep{mitra2024orca}, MathInstruct~\citep{yue2023mammoth}, MetaMath~\cite{yu2023metamath}, and from code category, including Magicoder~\citep{wei2024magicoder}, WizardCoder~\citep{luo2023wizardcoder}, and GlaiveCodeAssistant~\citep{GlaiveCodeAssistant2023}. After that, we leverage OpenAI models, GPT-4o~\citep{gpt4o} and GPT-4o-mini~\citep{gpt4omini}, to further refine the responses of the prompts from these datasets to enhance the quality of our SFT dataset. Finally, we conduct data decontamination to make sure our dataset does not contain the prompts from all benchmark test datasets.

\section{Results}
\label{sec:results}
In this section, we present a comprehensive evaluation of the NVLM-1.0 model family across a wide range of benchmarks to assess their multimodal capabilities, comparing them to other leading open-access and proprietary multimodal LLMs. Additionally, we evaluate the NVLM-1.0 models and other top open-access multimodal LLMs on key text-only benchmarks, demonstrating either no degradation or even improvements in the text-only performance of the NVLM-1.0 models, in sharp contrast to the significant degradation observed in other open-access multimodal LLMs.

\subsection{Benchmarks}
We first introduce the vision-language and text-only benchmarks used in this work. 
Following previous frontier-class multimodal LLMs~\citep[e.g.,][]{gpt4o, claude3.5sonnet, 2024internvl2}, we evaluate NVLM on nine vision-language benchmarks, focusing on multimodal reasoning, math reasoning in visual context, natural image understanding, scene-text reading, chart understanding, document understanding, real-world perception, and OCR capabilities:
\begin{enumerate}[leftmargin=2.5em]
{
\item[$\diamond$]
MMMU~\citep{yue2023mmmu} is one of the most popular multimodal reasoning benchmarks, covering multidisciplinary college-level problems. We do evaluations on both validation and test sets.
%
\item[$\diamond$]
MathVista~\citep{lu2024mathvista} is a math reasoning benchmark that covers a variety of mathematical problems, e.g., geometry, function plot, table/chart related arithmetic, in visual contexts. We perform the evaluation on its testmini set.
%
\item[$\diamond$]
VQAv2~\citep{goyal2017making} is a natural image understanding benchmark. We evaluate NVLM models on the test-dev set.
%
\item[$\diamond$]
AI2D~\citep{kembhavi2016diagram} is a multimodal reasoning dataset with Grade School Science diagrams.We evaluate the test set using two evaluation settings from VLMEvalKit~\cite{duan2024vlmevalkit} (see Appendix \ref{fig:ai2d_illustration} for examples). 
In the first setting (``test''), the text in the image is replaced with letter options from the answer choices. In the second setting (``test\_no\_mask''), the text in the image is replaced with both the letter option and the corresponding value of the answer choices, which we refer to as \emph{no\_mask}. Note that the first setting is used as the default metric unless \emph{no\_mask} is explicitly stated.
%
\item[$\diamond$]
TextVQA~\citep{singh2019towards} is a scene-text reading benchmark that includes various text-reading problems from natural images. We condudt evaluation on its validation set.
%
\item[$\diamond$]
ChartQA~\citep{masry2022chartqa} is a chart understanding benchmark that involves visual and logical reasoning. We perform evaluation on its test set.
%
\item[$\diamond$]
DocVQA~\citep{mathew2021docvqa} is dataset for VQA on document images. We do evaluation on its test set.
%
\item[$\diamond$]
RealWordQA~\citep{xai2024Grok1-5} is a benchmark focused on physical world perception and understanding.
\vspace{-.05cm}
\item[$\diamond$]
OCRBench~\citep{liu2024OCRBench} is a comprehensive benchmark created to evaluate the OCR capabilities of multimodal LLMs. It consists of five components: text recognition in images, scene text-centric VQA, document-oriented VQA, key information extraction, and handwritten mathematical expression recognition.
}
\end{enumerate}

To assess the degradation of text-only performance during multimodal training, we evaluate and compare the multimodal LLMs against their corresponding text-only LLM backbones across four key benchmarks, focusing on multidisciplinary knowledge reasoning, math reasoning, and coding capabilities:
\begin{enumerate}[leftmargin=2.5em]
{
\item[$\circ$]
MMLU~\citep{hendrycks2020mmlu} is a multidisciplinary benchmark that covers 57 subjects, including elementary mathematics, U.S. history, computer science, law, and more
%
\item[$\circ$]
GSM8K~\citep{cobbe2021gsm8k} is a benchmark consisting of grade school math word problems.
%
\item[$\circ$]
MATH~\citep{hendrycks2021math} is a math reasoning benchmark that covers math problems ranging across 5 levels of difficulty and 7 sub-disciplines.
%
\item[$\circ$]
HumanEval~\citep{chen2021evaluating} is a coding benchmark that measures functional correctness for synthesizing programs from docstrings.
}
\end{enumerate}

\subsection{Baseline Models}
We compare our models to leading proprietary and open-access multimodal LLMs. The state-of-the-art (SOTA) proprietary models include GPT-4o~\citep{gpt4o}, Claude 3.5~\citep{claude3.5sonnet}, Gemini Pro 1.5~\citep{gemini1-5}, and Grok-2~\citep{xai2024Grok2}.
The SOTA open-access models include  InternVL-2-Llama3-76B~\citep{2024internvl2-llama3-76B}, InternVL 2-Pro~\citep{2024internvl2}, LLaVA-OneVision 72B~\citep{li2024llava-onevision}, Llama 3-V 70B and 405B~\citep{llama3-1}. Note that the model weights of top-performing InternVL 2-Pro (size unspecified) and Llama 3-V have not yet been made open-access.
We optimize and evaluate the following NVLM-1.0 models: \emph{i)} decoder-only NVLM-D$_{\text{ 1.0}}$ 72B, which process image tokens within the LLM decoder, 
\emph{ii)} cross-attention-based NVLM-X$_{\text{ 1.0}}$ 72B, which handle image tokens through $X$-attention layers, 
and \emph{iii)} hybrid NVLM-H$_{\text{ 1.0}}$ 72B, which process global thumbnail image tokens using self-attention layers in the LLM decoder, and regular tile image tokens using $X$-attention layers.

Given limited computational resources and the goal of building a frontier-class multimodal LLM, we used the smaller 34B models for faster ablation studies and iterations, without focusing on careful checkpoint selection or hyperparameter optimization.
We include the unoptimized 34B results in Appendix~\ref{appendix:nvlm-34b-results} for reference purpose.
Note that, although our 34B results are not optimized, the 34B NVLM models still significantly outperform other models, including VILA-1.5 40B~\citep{lin2024vila} and Cambrian-1 34B~\citep{tong2024cambrian}.

\begin{table*}[t!]
\caption{\footnotesize
Evaluation on vision-language and text-only benchmarks. For vision-language benchmarks, all baseline model results are sourced from official reports and the benchmark hosts. 
For open multimodal LLMs, we list the models that were open-access at the time of this report's publication and mark \textbf{*} for {\textcolor{gray}{\textbf{models}}} not yet open-access.
We \textbf{highlight} the highest score for each benchmark in both the proprietary and open-access categories.
\textbf{Text-only Avg. 4} represents the average accuracy degradation~({\color{red}--}) or improvement~({\color{emerald}+}) of the multimodal LLM compared to its backbone LLM on text-only benchmarks after multimodal training, measured across four key benchmarks: MMLU, GSM8K, MATH, and HumanEval~(see Table~\ref{tab:text-only} for full results).
}
\centering
\begin{adjustbox}{width={1.01\textwidth}}
\label{tab:main_result}
\begin{tabular}{lrccccccccc}
\toprule
\multirow{2}{4em}{Tasks} 
 & MMMU  & MathVista & VQAv2 & AI2D   &  TextVQA  & ChartQA & DocVQA & {Real-} & OCR- & \textbf{Text-only}  \\
 & test~/~val  & testmini & test-dev  & test~/~\emph{no\_mask} & val & test & test & WorldQA & Bench  &  \textbf{Avg. 4}  \\
\midrule
%
\hspace{-.15cm}\textbf{\emph{Proprietary}}  \\
\midrule  
GPT-4V~\cite{gpt4v} & \textbf{56.1}~/~56.8 & 49.9 & 77.2 & 78.2 & 78.0 & 78.5 & 88.4 & 61.4  & 645  & -  \\
GPT-4-Turbo~\citep{gpt4turbo} & -~~/~63.1 & 58.1 & - & 89.4 & - & 78.1 & 87.2 & - & 678 & - \\
GPT-4o~\citep{gpt4o} & -~~/~\textbf{69.1} & 63.8 & - & 94.2 & - & 85.7 & 92.8 & - & 736 &  - \\
\midrule
Claude 3 Sonnet~\citep{claude3}  & -~~/~53.1 & 47.9 & - & 88.7 & - & 81.1 & 89.5 & 51.9  & 646 & - \\
Claude 3 Opus~\citep{claude3}  & -~~/~59.4  & 50.5 & - & 88.1 & - & 80.8 & 89.3 & 49.8 & 694 & - \\
Claude 3.5 Sonnet~\citep{claude3.5sonnet} & -~~/~68.3  & 67.7 & - & \textbf{94.7} & - & \textbf{90.8} & \textbf{95.2} & -  &  \textbf{788} & - \\
\midrule
Gemini Pro 1.0~\citep{gemini1-0} & -~~/~47.9  &   45.2 & 71.2  & 73.9 & 74.6 &  74.1 & 88.1  &  - & 659 & - \\
Gemini Ultra 1.0~\citep{gemini1-0}  & -~~/~59.4 &  53.0 & 77.8 & 79.5 & \textbf{82.3} & 80.8 & 90.9 & - & -  & -  \\
Gemini Pro 1.5~\citep{gemini1-5} &  -~~/~58.5 &  52.1 & 80.2  & 80.3 & 73.5 & 81.3 & 86.5 & 67.5 & - &  - \\
Gemini Pro 1.5~(Aug 2024) &  -~~/~62.2 &  63.9 & \textbf{80.2}  & 94.4 & 78.7 & 87.2 & 93.1 & \textbf{70.4} & 754 &  - \\
\midrule
Grok-1.5V~\citep{xai2024Grok1-5} & -~~/~53.6 &  52.8 & -  & 88.3 & 78.1 & 76.1 & 85.6 &  68.7  & - & - \\
Grok-2~\citep{xai2024Grok2} & -~~/~66.1  & \textbf{69.0} & - & - & - & - & 93.6 &  - & - & - \\
\midrule
\hspace{-.1cm}\emph{Others} \\
QWen-VL-MAX  &  46.8~/~51.4  & 51.0  &  78.8  & 79.3 & 79.5  & 79.8 & 93.1 & -  & 723  & - \\
Adept Fuyu-Heavy~\citep{adept2024-fuyu-heavy} & -~~/~48.3 & - & 77.8 & 	81.2  & - & 75.4 & -  & - & - & - \\
\midrule
\midrule
\hspace{-.15cm}\textbf{\emph{Open-access}} \\
\midrule  
LLaVA-Next 34B~\citep{liu2024llavanext}  & 44.7~/~51.1 & 46.5 & - &  - &  69.5 & - & - & -  & 574 & - \\ 
VILA-1.5 40B~\citep{lin2024vila} & 46.9~/~51.9 & - & 84.3 &  - & - & - & - & - & - & {\color{red} -- 6.9} \\
Cambrian-1 34B~\citep{tong2024cambrian}  & ~-~~/~49.7  & 53.2 & - & 79.7 & 76.7 & 75.6  & 75.5  & 67.8  &  600
& - \\
LLaVA-OneVision 72B~\citep{li2024llava-onevision}  &  -~~/~56.8  &  67.5  & -  & 85.6 & -  & 83.7  & 91.3  & -  & -  & {\color{red} -- 6.3}  \\
\midrule
InternVL-1.2 40B~\citep{chen2024internvl} & -~~/~51.6 & 47.7 & -  & 79.0 & 72.5 & 68.0 & 57.7 & 67.5 & 569 & - \\
InternVL-1.5 26B~\citep{chen2024internvl1-5}   & -~~/~45.2  & 53.5  & - &  80.7 & 80.6 & 83.8 & 90.9 &  66.0 & 724 & - \\
InternVL-2 40B~\citep{2024internvl2} & -~~/~53.9 & 63.7 & - & 87.1 & 83.0 & 86.2 & 93.9 & 71.8 & 837 & - \\
InternVL-2-Llama3-76B &  -~~/~55.2 &  65.5 & - & 87.6 / 94.8 & \textbf{84.4} & \textbf{88.4} & 94.1 & \textbf{72.2} & 839 &  {\color{red} -- 6.7} \\
*\textcolor{gray}{InternVL-2-Pro}~\citep{2024internvl2} & -~~/~58.9 &  66.3  &  - &  87.3~/~\textbf{96.0}  & - & 87.1 & \textbf{95.1} & - &  837 & - \\
\midrule
*\textcolor{gray}{Llama 3-V 70B}~\citep{dubey2024llama}  &  -~~/~60.6 & - &  79.1 &  93.0 & 83.4 & 83.2 & 92.2  & - & - & 0 \\
*\textcolor{gray}{Llama 3-V 405B}~\citep{dubey2024llama}  &  -~~/~{64.5} & - &  80.2 & 94.1 & 84.8 & 85.8 &  92.6 & - & - & 0 \\
\midrule
NVLM-D$_{\text{ 1.0}}$ 72B 
& \textbf{54.6}~/~\textbf{59.7} & 65.2 & \textbf{85.4} & 85.2 / 94.2  & 82.1 & 86.0 & 92.6 & 69.7  & \textbf{853}  &  {\color{emerald}\textbf{+ 4.3}}   \\  
NVLM-X$_{\text{ 1.0}}$ 72B 
& 53.6~/~57.4 & 64.6 & 85.2 & 84.2 / 93.6 & 80.2 & 82.9 & 82.9 & 66.1 & 828 & {\color{emerald}+ 2.5}  \\ 
NVLM-H$_{\text{ 1.0}}$ 72B & 53.0~/~\textbf{60.2} & \textbf{66.6} & 85.2 & 83.8~/~93.3 & 80.3 & 83.3 & 83.1 & 66.0 & 831 &   {\color{emerald}+ 2.7} \\
\bottomrule
\end{tabular}
\end{adjustbox}
\end{table*}

\subsection{Main Results}
The main results are presented in Table~\ref{tab:main_result}, which includes outcomes from nine vision-language benchmarks and four text-only benchmarks. 
Our NVLM-1.0 72B models rival the leading proprietary models (e.g., GPT-4o) and open-access models, including LLaMA 3V (not yet publicly available) and InternVL 2.
Specifically, the following observations can be drawn from Table~\ref{tab:main_result}:
\begin{enumerate}[leftmargin=2.5em]
{
\item[$\bullet$]
NVLM-D$_{1.0}$ 72B achieves the highest scores on OCRBench (853) and VQAv2 (85.4) among all leading proprietary and open-access models. Its MMMU score (59.7) also significantly surpasses all leading open-access models at the time of this report's publication, including LLaVA-OneVision 72B (56.8)~\citep{li2024llava-onevision} and InternVL-2-Llama3-76B (55.2)~\citep{2024internvl2-llama3-76B}.
On AI2D, TextVQA, ChartQA, and DocVQA, it performs only slightly worse than the best-performing InternVL-2-Llama3-76B, matches very strong GPT-4o~\citep{gpt4o}, and significantly outperforms other leading open-access models, including Cambrian-1~\citep{tong2024cambrian} and LLaVA-OneVision 72B~\citep{li2024llava-onevision}.
\item[$\bullet$]
NVLM-H$_{1.0}$ 72B achieves the highest MMMU~(Val) score~(60.2) among all multimodal LLMs that are open-access at the time of this report's publication.  It also achieves the best MathVista score~(66.6) within NVLM-1.0 family, which already outperforms many very strong models including GPT-4o~\citep{gpt4o}, Gemini Pro 1.5~(Aug 2024)~\citep{gemini1-5}, InternVL-2-Pro~\citep{2024internvl2}.
This demonstrate its superb multimodal reasoning capability.
\item[$\bullet$]
NVLM-X$_{1.0}$ 72B also achieves frontier-class results and stands as the best-in-class cross-attention-based multimodal LLMs, rivaling the yet-to-be-released Llama 3-V 70B~\citep{llama3-1}.
One notable advantage of NVLM-X$_{1.0}$ is its significantly faster training and inference speeds compared to its decoder-only counterpart, as demonstrated in Table~\ref{tab:traning_speed_nvlm}.
\item[$\bullet$]
Open-access multimodal LLMs, such as LLaVA-OneVision 72B and InternVL-2-Llama3-76B, show significant performance degradation on text-only tasks after multimodal training. In contrast, our NVLM-1.0 models exhibit even improved text-only performance, thanks to the inclusion of high-quality text-only SFT data. This demonstrates that unfreezing the LLM backbone during multimodal SFT does not compromise text performance, as long as high-quality text alignment data is incorporated.
}
\end{enumerate}

\begin{table*}[t!]
\caption{\footnotesize
Evaluation on text benchmarks: MMLU, GSM8K, MATH and HumanEval.
For leading proprietary models, information about potential text performance degradation during multimodal training has not been disclosed.
The model weights of *\textcolor{gray}{LLaMA 3-V} had not been released at the time of this report.
Text-only Avg. 4 represents the average accuracy degradation~({\color{red}--}) or improvement~({\color{emerald}+}) of the multimodal LLM compared to its backbone LLM on text benchmarks after multimodal training.
}
\vspace{-.2em}
\centering
\begin{adjustbox}{width={\textwidth}}
\label{tab:text-only}
\begin{tabular}{lccccccc}
\toprule
\multirow{2}{4em}{Tasks}   & Backbone  & \multirow{2}{4em}{MMLU}  & \multirow{2}{4em}{GSM8K} & \multirow{2}{4em}{MATH}  &  \multirow{2}{4em}{HumanEval}  &  Avg.   &  Text-only  \\
& LLM   & & & & & Accuracy & Avg. 4 \\
\midrule
\hspace{-.15cm}\textbf{\emph{Proprietary}} & & & & & 
\\
GPT-4o~\citep{gpt4o}  & N/A  & 88.7 & - &  76.6 & 90.2 &  -  & unknown  \\
Gemini Pro 1.5~(Aug 2024)~\citep{gemini1-5} & N/A & 85.9 & 90.8 & 67.7 & 84.1  & 
 82.1 & unknown \\
Claude 3.5 Sonnet~\citep{claude3.5sonnet} & N/A  & 88.7 & 96.4  & 71.1  & 92.0 & \textbf{87.0}  & unknown
\\
\midrule
\hspace{-.15cm}\textbf{\emph{Open LLM}}\\
{\color{teal}\emph{(a)}}~Nous-Hermes-2-Yi-34B~\citep{Nous-Hermes-2-Yi-34B} & N/A & 75.5 & 78.6 & 21.8 & 43.3 &  54.8 & N/A \\
{\color{violet}\emph{(b)}}~Qwen2-72B-Instruct~\citep{qwen2} & N/A & 82.3 & 91.1 & 59.7 & 86.0 & 79.8  & N/A \\
%
{\color{cyan}\emph{(c)}}~Llama-3-70B-Instruct~\citep{dubey2024llama} & N/A  &  82.0 & 93.0 & 51.0 & 81.7 & 76.6 & N/A \\
{\color{blue}\emph{(d)}}~Llama-3.1-70B-Instruct~\citep{dubey2024llama} & N/A  &  83.6 & 95.1 & 68.0 & 80.5 & 81.8 & N/A \\
{\color{Blue}\emph{(e)}}~Llama-3.1-405B-Instruct~\citep{dubey2024llama} & N/A  &  87.3 & 96.8 & 73.8 & 89.0 & 86.7 & N/A 
\\
\midrule  
\hspace{-.15cm}\textbf{\emph{Open Multimodal LLM}}\\
VILA-1.5 40B~\citep{lin2024vila} & {\color{teal}\emph{(a)}} & 73.3 & 67.5 & 16.8 & 34.1 &  47.9 & {\color{red}-- 6.9} \\
LLaVA-OneVision 72B~\citep{liu2024llavanext} & {\color{violet}\emph{(b)}}  & 80.6 & 89.9 & 49.2 & 74.4  & 73.5 &   {\color{red}-- 6.3} \\
InternVL-2-Llama3-76B~\citep{2024internvl2} & {\color{cyan}\emph{(c)}}  & 78.5 & 87.1 & 42.5 & 71.3 & 69.9 &  {\color{red}\ -- 6.7 } \\
*\textcolor{gray}{Llama 3-V 70B}~\citep{dubey2024llama}  &
{\color{blue}\emph{(d)}} & 83.6 & 95.1 & 68.0 & 80.5 & 81.8 &  0 \\
*\textcolor{gray}{Llama 3-V 405B}~\citep{dubey2024llama}  &
{\color{Blue}\emph{(e)}}&  87.3 & 96.8 & 73.8 & 89.0 & 86.7 &  0
\\
\midrule
%
%
NVLM-D$_{\text{ 1.0}}$  72B &  {\color{violet}\emph{(b)}} & 82.0 & 92.9 & 73.1  &  88.4 & \textbf{84.1} & {\textcolor{emerald}{\textbf{+ 4.3}}} \\ 
NVLM-X$_{\text{ 1.0}}$  72B & {\color{violet}\emph{(b)}}    & 81.4 & 91.8 & 70.6 & 85.2 & 82.3 & {\textcolor{emerald}{+ 2.5}} \\ 
NVLM-H$_{\text{ 1.0}}$  72B & {\color{violet}\emph{(b)}}    & 80.4 & 91.5 & 71.4 & 86.6 & 82.5 & {\textcolor{emerald}{+ 2.7}} \\ 
\bottomrule
\end{tabular}
\end{adjustbox}
\end{table*}

\subsection{Text-only Performance}
We present detailed results of text-only performance for our NVLM models, along with leading proprietary and open-access multimodal LLMs, in Table~\ref{tab:text-only}.
It can be observed that all open-access models experience a significant drop in accuracy compared to their LLM backbones. For instance, VILA-1.5 40B sees a notable decrease of 6.9 points, from 54.8 to 47.9. Similarly, the average accuracy of LLaVA-OneVision 72B and InternVL-2-Llama3-76B drops by 6.3 and 6.9 points, respectively.
Llama 3-V experiences no degradation in text-only performance because the LLM is frozen during multimodal training.
However, as we will demonstrate in \S~\ref{subsec:forzen_vs_unfrozen_LLM}, this frozen LLM strategy may lead to an unnecessary trade-off in vision-language performance.

By incorporating high-quality text SFT data, both the NVLM-1.0 72B models achieve higher average accuracy than their respective LLM backbone, Qwen2-72B-Instruct~\citep{qwen2}. For example, the average accuracy across four benchmarks increases from 79.8 to 84.1 for the NVLM-D$_{1.0}$ 72B model.
It is particularly interesting that the math capabilities of the NVLM-1.0 models improve significantly compared to its text-only backbone. We attribute this to the high-quality text-only SFT data and the substantial amount of multimodal math data included in our training blend, which enhances math reasoning skills overall, regardless of the modality.

\begin{table}[!t]
\centering
\vspace{-.2cm}
\caption{\footnotesize
Impact on vision-language performance with a frozen vs. unfrozen LLM backbone during multimodal SFT for the cross-attention model NVLM-X.}
\vspace{.3em}
\begin{adjustbox}{width={1\textwidth}}
\begin{tabular}{lccccccccl}
\toprule
\multirow{2}{4em}{Tasks} 
 & MMMU  & MathVista & VQAv2 & AI2D   &  TextVQA  & ChartQA & DocVQA & {RealWorld-} & OCR-   \\
 & test~/~val  & testmini & test-dev  & test & val & test & test & QA & Bench  \\
\midrule
NVLM-X 34B (frozen) & 43.2~/~51.6    & 51.8   & 83.8   & 72.4 & 72.4    & 74.4    & 73.2   & 63.4     & 696     \\ 
NVLM-X 34B  & 47.2~/~54.0     & 59.2 & 84.5      & 79.6 & 78.2    & 79.4    & 79.2   & 64.8        & 802    \vspace{.1cm} 
\\
NVLM-X 72B (frozen) & 50.6~/~54.4   & 60.6     & 85.3 & 76.2   & 76.2   & 76.2 & 76.4     & 65.3 & 722  \\
NVLM-X$_{\text{ 1.0}}$ 72B 
& 53.6~/~57.4 & 64.6 & 85.2 & 84.2 & 80.2 & 82.9 & 82.9 & 66.1 & 828     \\
\bottomrule
\end{tabular}
\end{adjustbox}
\label{tab:frozen_unfrozen_LLM_study}
\end{table}

\subsection{Frozen versus Unfrozen LLM during Mutimodal SFT}
\label{subsec:forzen_vs_unfrozen_LLM}
In this subsection, we compare two methods for maintaining text-only performance in the cross-attention-based NVLM-X: \emph{i)} Freezing the LLM during multimodal SFT training, which ensures no degradation in text performance due to the gated $X$-attention layers, and \emph{ii)} our default approach, which incorporates a high-quality text-only dataset during multimodal SFT training.
It is important to note that freezing the LLM for decoder-only multimodal model during SFT leads to poor results on vision-language tasks (as demonstrated in a similar study by \citep{lin2024vila}), due to the very limited capacity of the MLP projector module.

In Table~\ref{tab:frozen_unfrozen_LLM_study}, it can be seen that freezing the LLM yields reasonably good results. Notably, accuracy scales well as the model size increases from 34B to 72B, reaffirming findings from the original Flamingo study, which also froze the LLM during multimodal training. However, compared to the unfrozen setting, freezing the LLM still results in a moderate performance drop on vision-language tasks. For example, NVLM-X$_{1.0}$ 72B with a frozen LLM performs only comparably to the smaller NVLM-X-34B, showing slight improvements on multimodal reasoning tasks~(MMMU and MathVista) and natural image understanding tasks (VQAv2 and RealWorldQA), while falling behind on OCR-related tasks (AI2D, TextVQA, ChartQA, DocVQA, and OCRBench).

\section{Conclusion}
\label{sec:conclusion}
We introduce NVLM-1.0, a family of frontier multimodal large language models that achieve state-of-the-art results on vision-language tasks, rivaling leading multimodal LLMs, without compromising text-only performance during multimodal training.
Furthermore, we provide key insights on architecture design, tile-based dynamic high-resolution input, multimodal training data curation, and how to achieve production-grade multimodality with even improved text-only performance after multimodal training.
%
We open-source the model weights and training code to the community. Please refer to the updated information at \url{https://nvlm-project.github.io/}.

\section{Acknowledgement}
We would like to express our sincere gratitude to Jared Casper, Peter Dykas, and Mike Chrzanowski for their support with Megatron-LM. 
Our thanks also go to Lukas Voegtle and Philipp Fischer for the implementation and debugging efforts with their Megatron-Energon dataloader, which is released early. We are also grateful to Hongxu Yin, Timo Roman, Karan Sapra and Valerii Iakovlev for their help and support. 


{
\small
\bibliography{paper}
\bibliographystyle{neurips_2024}

}

\newpage


\appendix

\section*{\LARGE Appendix}

\section{Qualitative Examples from the NVLM-1.0-D 72B Model}
\label{appendix:more_examples}

\vspace{-0.5cm}
\begin{figure}[h!]
\centerline{\includegraphics[width=1\linewidth]{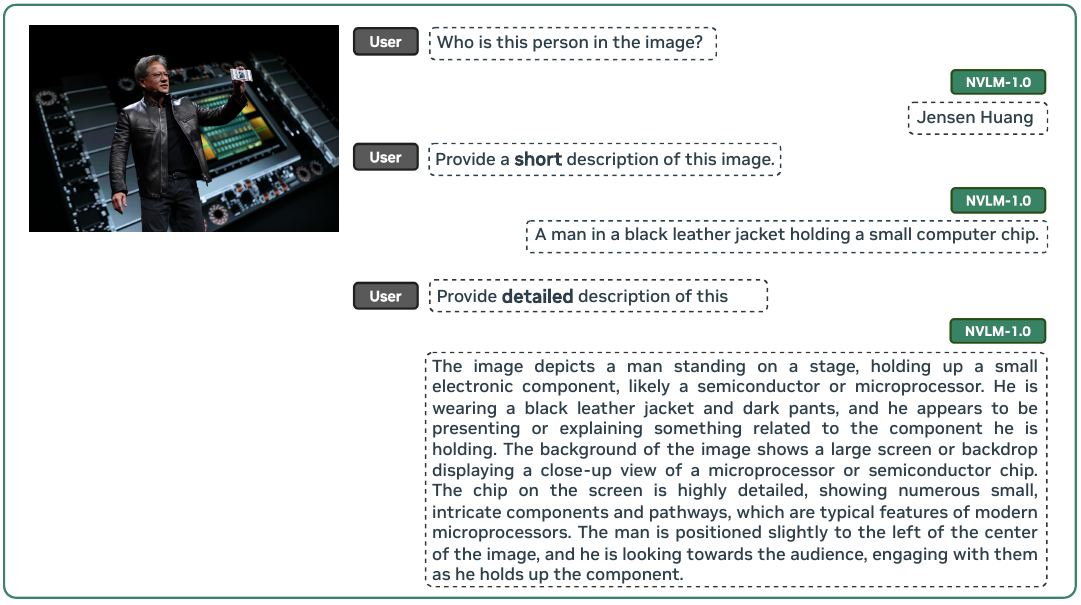}}
\vspace{-0.1cm}
\caption{\footnotesize
NVLM-1.0-D 72B model demonstrates good instruction-following capability. Based on the instructions, it appropriately controls the target generation's length. It can also generate a very high-quality, detailed description of the provided image.
}
\label{fig:qualitative_jensen}
\end{figure}

\vspace{-0.2cm}
\begin{figure}[h!]
\centerline{\includegraphics[width=1.0\linewidth]{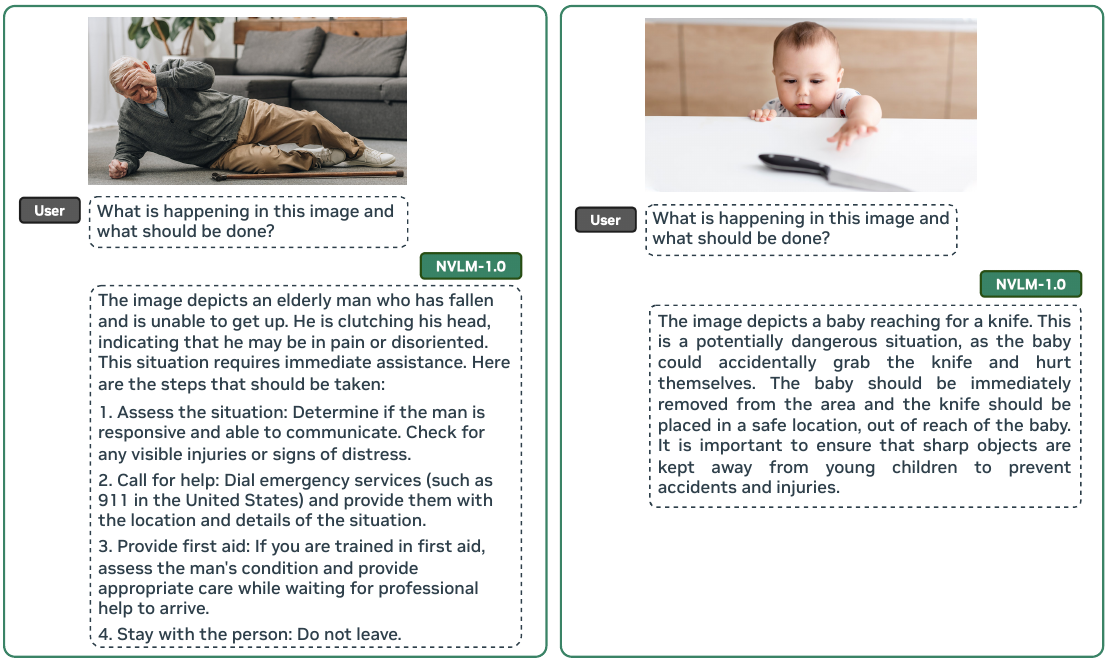}}
\vspace{-.1cm}
\caption{\footnotesize Demonstration of NVLM-1.0-D 72B model's strong scene understanding capability. It has common-sense knowledge to recognize potential dangers or accidents and correctly suggests what immediate actions should be taken. 
}
\label{fig:qualitative_scene}
\end{figure}

\vspace{-2.5cm}
\begin{figure}[h!]
\centerline{\includegraphics[width=1\linewidth]{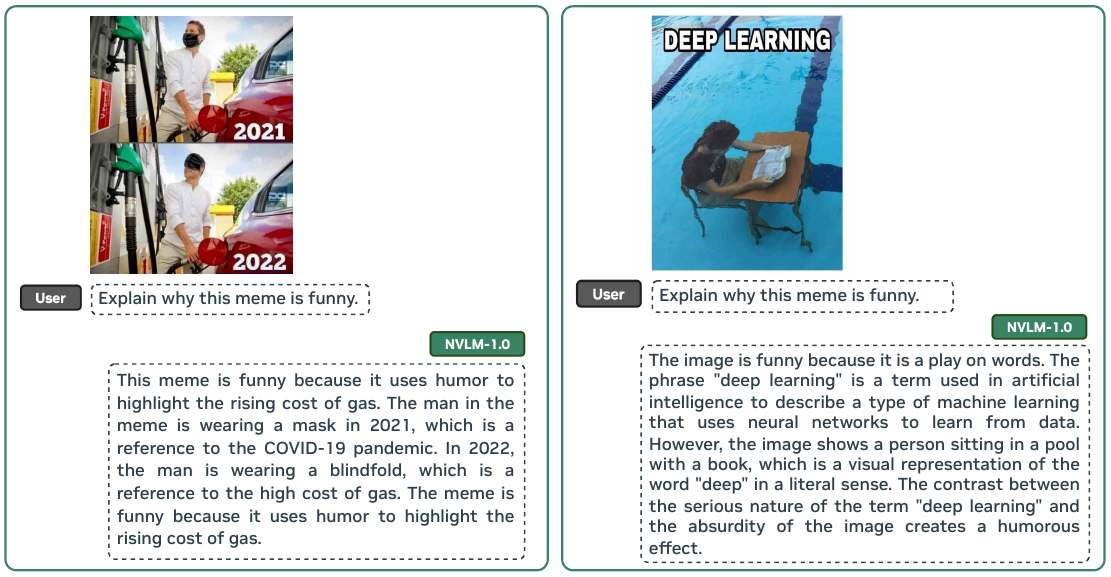}}
\vspace{-0.1cm}
\caption{\footnotesize
More examples of NVLM-1.0-D 72B model's ability to understand memes, which is a challenging task that requires an understanding of humor and knowledge of important social trends, context, or events.
}
\label{fig:qualitative_meme}
\end{figure}

\begin{figure}[h!]
\centerline{\includegraphics[width=1.\linewidth]{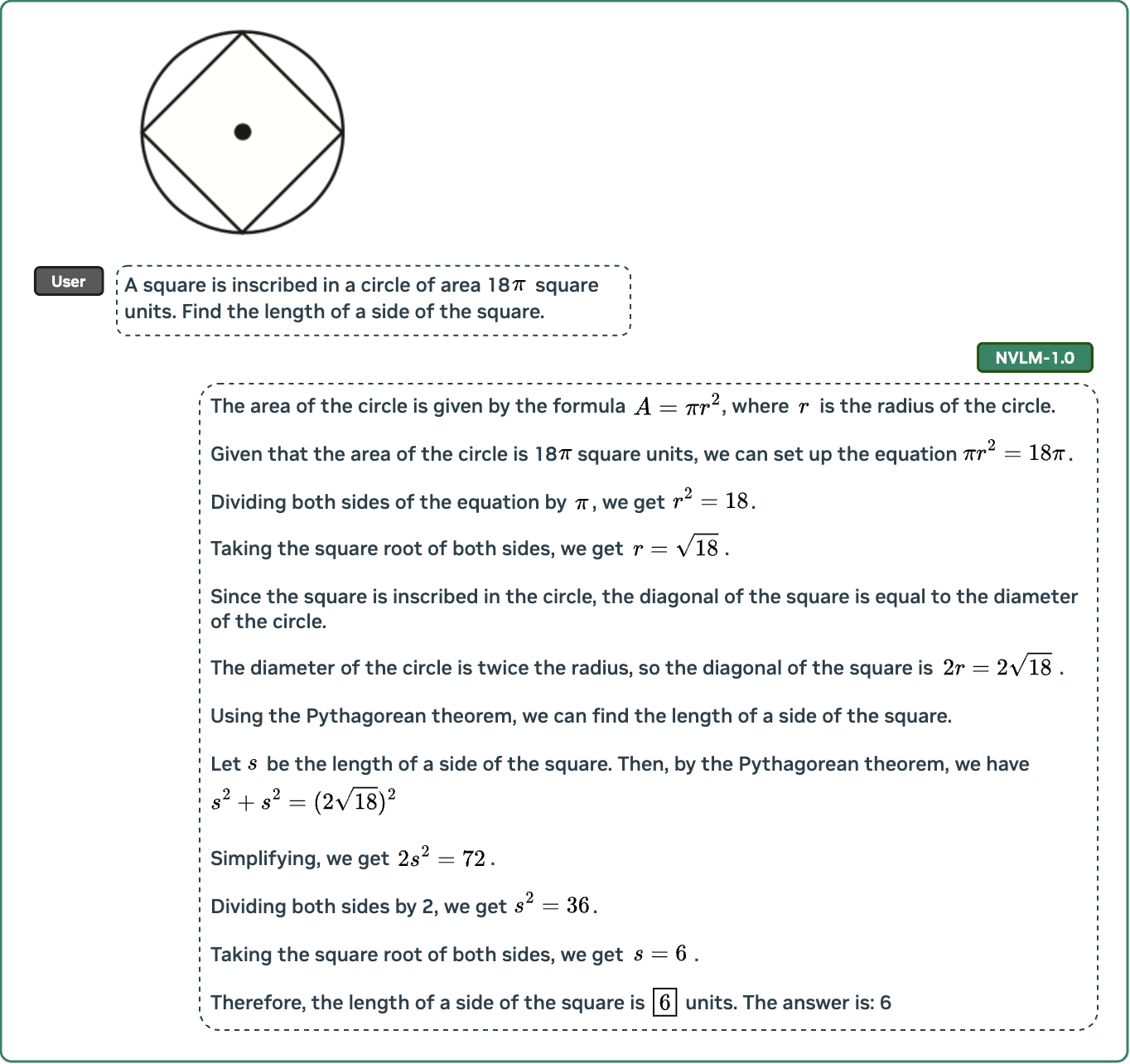}}
\vspace{-0.1cm}
\caption{\footnotesize NVLM-1.0-D 72B model can solve math questions by providing step-by-step mathematical reasoning. We render latex equations for readability. 
}
\label{fig:qualitative_math}
\end{figure}

\clearpage

\section{Training Hyperparameters}
\label{appendix:training_configuration}
We provide the pretraining hyperparameters in Table~\ref{tab:hyperparams_pretraining} and the SFT hyperparameters in Table~\ref{tab:hyperparams_sft}.

\begin{table}[!htbp]
\centering
\caption{Training hyper-parameters of NVLM models in the pretraining stage.}
\vspace{2pt}
\begin{adjustbox}{width={1\textwidth}}
\setlength{\tabcolsep}{20pt} 
\renewcommand{\arraystretch}{1.5} 
\begin{tabular}{lccc}
\toprule
Hyper-parameters                   & NVLM-D                         & NVLM-X        & NVLM-H                         \\  \midrule
Trainable weights                  & MLP                            & X-attn layers & X-attn layers \& MLP           \\
\# of gated cross-attention layers & N/A                            & 10            & 10                             \\
Global batch size                  & 2,048                          &  2048         & 2,048                          \\
Max learning rate                  & 1e-4                           &  1e-4         & 1e-4                           \\
Min learning rate                  & 2.5e-5                         &  1e-5         & 1e-5                           \\
Learning rate warmup steps         & 500                            &  1,000        & 1,000                          \\
Scheduler                          & cosine                         & cosine        & cosine                         \\
Optimizer                          & AdamW                          & AdamW         & AdamW                          \\
Optimizer config                   & $\beta_1=0.9$, $\beta_2=0.95$  &    $\beta_1=0.9$, $\beta_2=0.95$           & $\beta_1=0.9$, $\beta_2=0.98$  \\
Weight decay                       & 0.1                            &      0.05         & 0.05                           \\
Gradient clipping                  & 10                             &   1.0            & 1.0                            \\
Sequence length in the LLM decoder               & 512                            &    512           & 512                            \\
Downsampling of visual tokens             & 1024->256                            &   1024->256           &   1024->256   
\\
\# of visual token per tile             & 256                            & 256         & 256                           
\\
\# of tiles             &  1                            &   1           &   6+1     \\
Tensor parallelism                 & 8                              & 8             & 8                              \\
Pipeline parallelism               & 1                              & 1             & 1                              \\ 
\# of training steps               & 20K                            & 20K           & 20K                            \\ \bottomrule
\end{tabular}
\end{adjustbox}
\label{tab:hyperparams_pretraining}
\end{table}

\begin{table}[!htbp]
\centering
\caption{Training hyper-parameters of NVLM models in the SFT stage.}
\vspace{2pt}
\begin{adjustbox}{width={1\textwidth}}
\setlength{\tabcolsep}{10pt} 
\renewcommand{\arraystretch}{1.5} 
\begin{tabular}{lccc}
\toprule
Hyper-parameters                   & NVLM-D                         & NVLM-X               & NVLM-H                         \\ 
\midrule
Trainable weights                  & MLP \& LLM                     & X-attn layers \& LLM & X-attn layers \& MLP \& LLM    \\
\# of gated cross-attention layers & N/A                            & 10                   & 10                             \\
Global batch size                  & 128                            & 512 (34B), 256 (72B) & 256                            \\
Max learning rate                  & $2e^{-6}$                      & $1e^{-5}$            & $1e^{-5}$                      \\
Min learning rate                  & $2.5e^{-7}$                    & $1e^{-6}$ (34B), $1e^{-7}$ (72B) & $1e^{-7}$          \\
Learning rate warmup steps         & 1,000                          & 500                  & 1,000                          \\
Scheduler                          & cosine                         & cosine               & cosine                         \\
Optimizer                          & AdamW                          & AdamW                & AdamW                          \\
Optimizer config                   & $\beta_1=0.9$, $\beta_2=0.98$  &    $\beta_1=0.9$, $\beta_2=0.98$                  & $\beta_1=0.9$, $\beta_2=0.98$  \\
Weight decay                       & 0.1                            &      0.05                & 0.05                           \\
Gradient clipping                  & 10                             &     1.0                 & 1.0                            \\
Sequence length in the LLM decoder                & 3,200                          & 1,024                & 1,280                          \\
Downsampling of visual tokens             & 1024->256                            &   1024->256           &   1024->256   
\\
\# of visual token per tile             & 256                            & 256         & 256                           
\\
\# of tiles             &  6+1                            &   6+1           &   6+1     
\\
Tensor parallelism                 & 8                              & 8                    & 8                              \\
Pipeline parallelism               & 4                              & 1                    & 1                              \\ 
\# of training steps               & 40K                            & 20K                  & 40K                            \\ \bottomrule
\end{tabular}
\end{adjustbox}
\label{tab:hyperparams_sft}
\end{table}

\section{Perceiver Resampler in Flamingo Impacts OCR Performance}
\label{appendix:perceiver_in_flamingo}
\begin{figure}[h!]
    \centering
    \begin{subfigure}{0.47\linewidth}
        \centering
        \includegraphics[width=\linewidth]{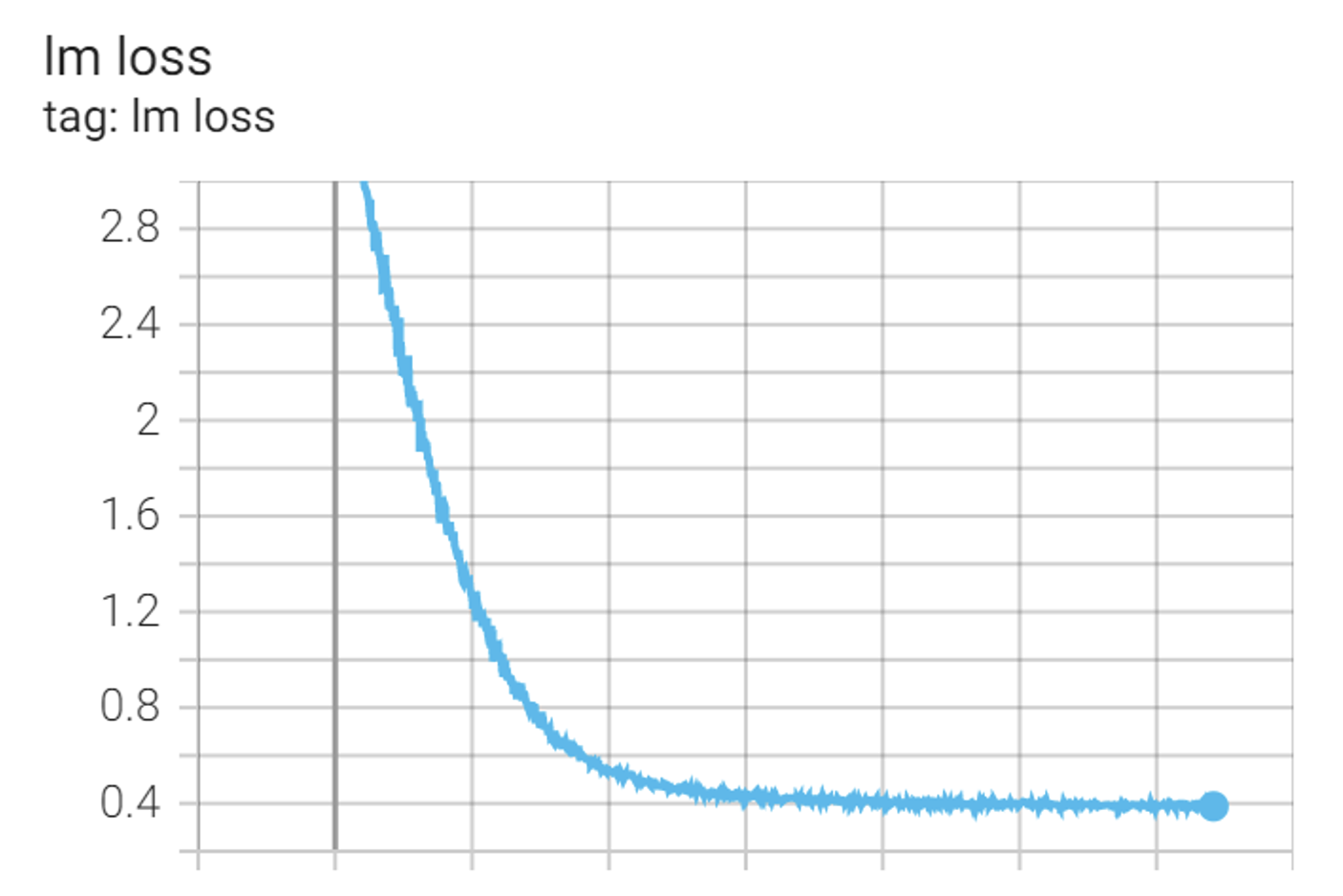}
        \caption{Training loss of Flamingo w/ perceiver.}
        \label{fig:llama3_1_instruct_8b}
    \end{subfigure}
    \hfill
    \begin{subfigure}{0.47\linewidth}
        \centering
        \includegraphics[width=\linewidth]{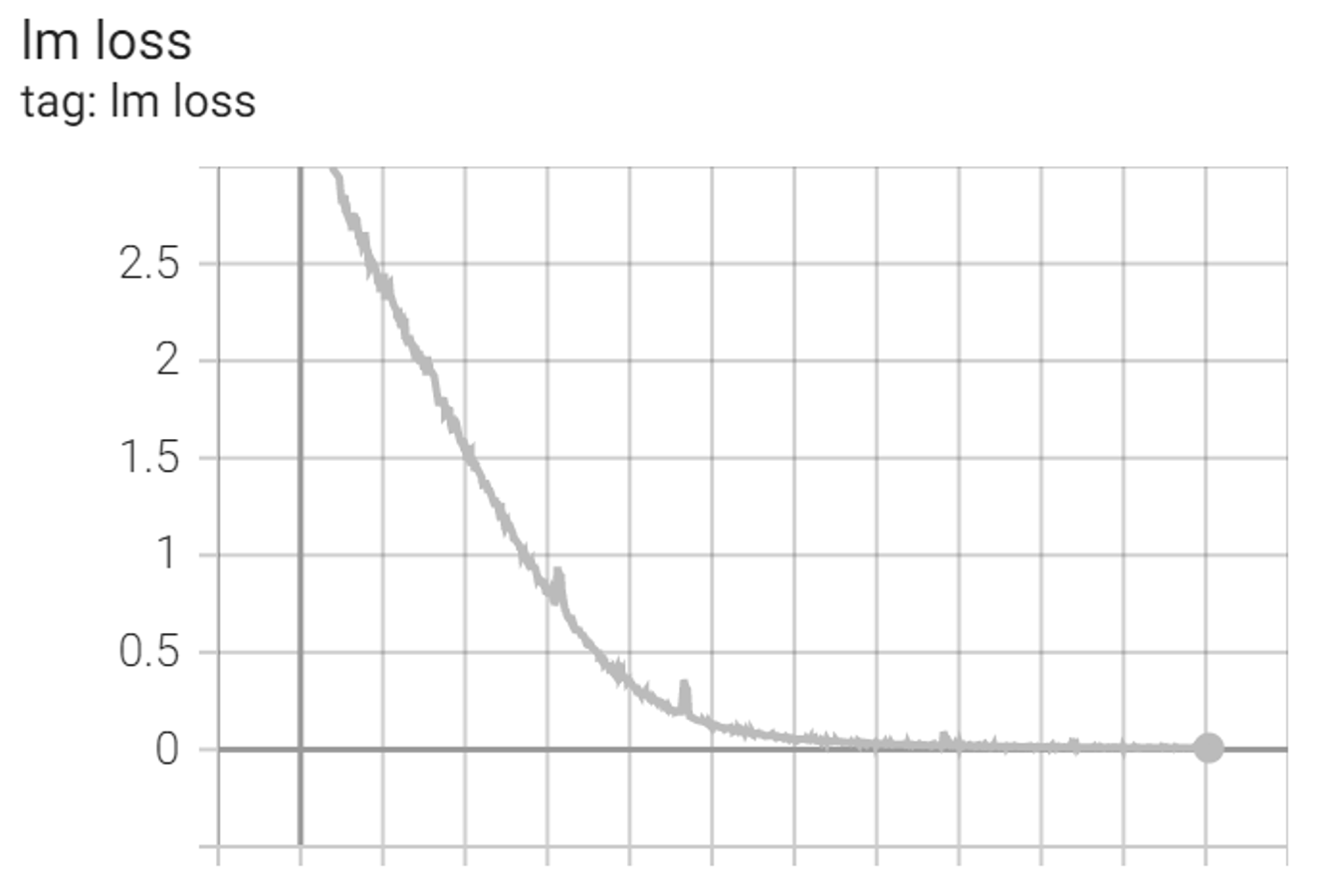}
        \caption{Training loss of ``Flamingo'' w/o perceiver.}
        \label{fig:llama3_1_instruct_70b}
    \end{subfigure}
    \caption{An overfitting experiment on the Flamingo models with and without the perceiver resampler on a document OCR dataset.}
    \label{fig:flamingo_perceiver_overfit}
\end{figure}

In this study, we utilize a pretrained Flamingo model~\citep{alayrac2022flamingo} from \citet{yang2023re}, built on a 1.3B LLM, and fine-tune it on an internal document OCR dataset consisting of 30K samples.
In Figure~\ref{fig:flamingo_perceiver_overfit}, we observe that the original Flamingo model, incorporating the \emph{perceiver resampler}, struggles to overfit this OCR dataset. For instance, even after numerous epochs, the training loss remains around 0.4.
However, when we remove the \emph{perceiver resampler} and only train the cross-attention layer, the loss decreases to 0 at the same iteration.
We hypothesize that the $X$-attention operation to the \emph{latent array} in Perceiver~\citep{jaegle2021perceiver} may shuffle the spatial information among the image patches, making it challenging for the subsequent cross-attention layer to disentangle.

\section{Evaluation Details of AI2D}
\label{appendix:evaluation}
\begin{figure}[h!]
\vspace{1cm}
    \centering
    \begin{subfigure}{0.47\linewidth}
        \centering
        \includegraphics[width=\linewidth]{figures/ai2d_with_mask.png}
        \caption{Evaluation Setting 1: With mask}
        \label{fig:ai2d_eval_mask}
    \end{subfigure}
    \hfill
    \begin{subfigure}{0.47\linewidth}
        \centering
        \includegraphics[width=\linewidth]{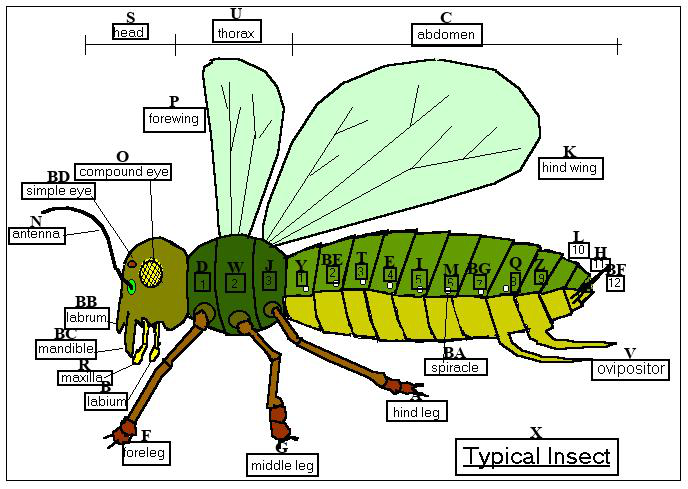}
        \caption{Evaluation Setting 2: With no mask}
        \label{fig:ai2d_eval_no_mask}
    \end{subfigure}
    \caption{Illustration of two AI2D evaluation settings adopted from VLMEvalKit using a test sample with the question ``Which is the leg closest to the head?''.}
    \label{fig:ai2d_illustration}
\end{figure}

We provide an illustration of two AI2D evaluation settings in Figure~\ref{fig:ai2d_eval_mask} and \ref{fig:ai2d_eval_no_mask}.

\section{Data Formats and ChatML Tamplate}
\label{appendix:chat_template}
\begin{figure}
\centerline{\includegraphics[width=0.9\linewidth]{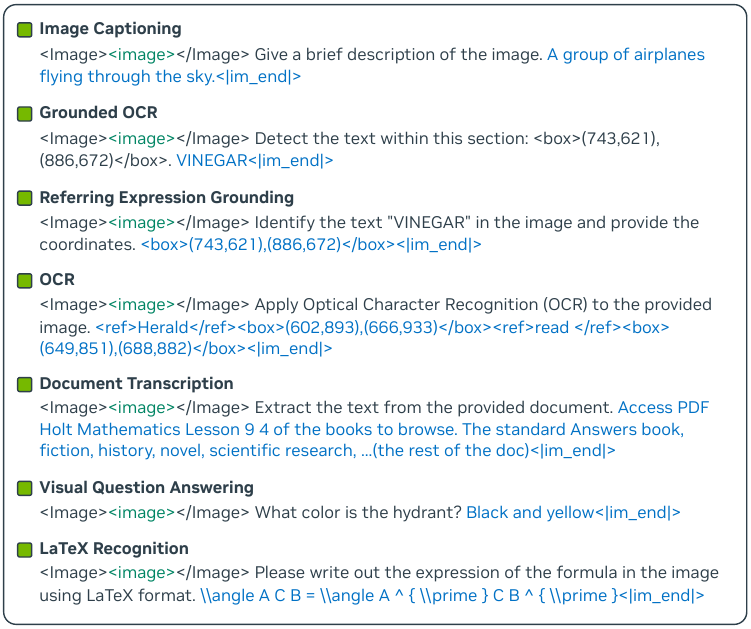}}
\caption{\footnotesize
Examples of training formats for various tasks used in pre-training. The emerald colored <image> tag indicates where to insert visual features. The blue colored text represents the ground truth associated with loss.
}
\label{fig:training-template}
\end{figure}

\begin{figure}
\centerline{\includegraphics[width=0.9\linewidth]{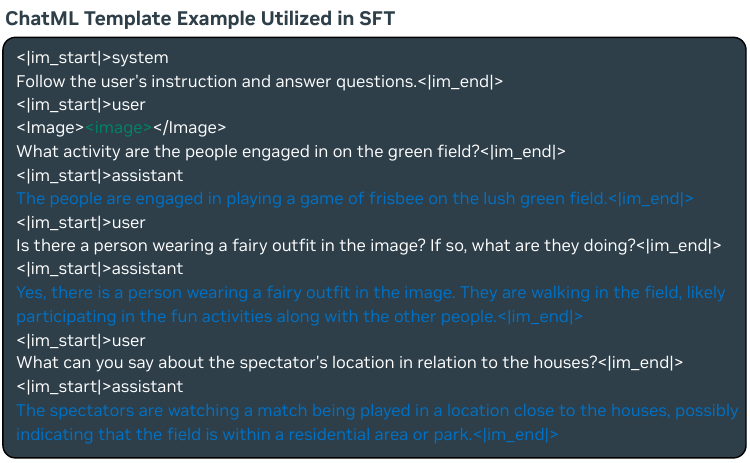}}
\caption{\footnotesize
An example of the ChatML template used in SFT. The emerald colored <image> tag indicates where to insert visual features. The blue colored text represents the ground truth associated with loss.
}
\label{fig:sft-template}
\end{figure}

We provide the examples of training data formats for various tasks in Figure~\ref{fig:training-template} and ChatML template used in SFT  in Figure~\ref{fig:sft-template}.


\newpage

\section{Unoptimized Results Using Yi-34B as the Backbone LLM}
\label{appendix:nvlm-34b-results}

We only utilized the smaller 34B models for faster ablation studies and iterations, without detailed checkpoint selection or hyperparameter optimization. The unoptimized 34B results are provided in the following Tables for reference.
Note that, although our 34B results are not optimized, the 34B NVLM models still significantly outperform other models, including VILA-1.5 40B~\citep{lin2024vila} and Cambrian-1 34B~\citep{tong2024cambrian}.

\begin{table*}[h!]
\caption{\footnotesize
Evaluation of 34B models on vision-language and text-only benchmarks.
\textbf{Text-only Avg. 4} represents the average accuracy degradation or improvement of the multimodal LLM compared to its backbone text-only LLM after multimodal training, measured across four text benchmarks: MMLU, GSM8K, MATH, and HumanEval~(see Table~\ref{tab:text-only} for full results).
}
\centering
\begin{adjustbox}{width={1.01\textwidth}}
\label{tab:34b_vl_result}
\begin{tabular}{lrccccccccc}
\toprule
\multirow{2}{4em}{Tasks} 
 & MMMU  & MathVista & VQAv2 & AI2D   &  TextVQA  & ChartQA & DocVQA & {Real-} & OCR- & \textbf{Text-only}  \\
 & test~/~val  & testmini & test-dev  & test~/~\emph{no\_mask} & val & test & test & WorldQA & Bench  &  \textbf{Avg. 4}  \\
\midrule
NVLM-D 34B  
& 48.7~/~52.1 &  \textbf{59.9} & 84.3 & \textbf{82.6} / \textbf{93.0} & \textbf{80.0} & \textbf{84.0} & \textbf{89.1} &  \textbf{67.3} & 819  & \textbf{+11.8} \\
NVLM-X 34B & 47.2~/~\textbf{54.0}     & 59.2 & 84.5      & 79.6 / 91.2 & 78.2    & 79.4    & 79.2   & 64.8        & 802  &  \textbf{+11.2} \\ 
NVLM-H 34B & 46.3~/~53.0 &  59.4  & \textbf{84.8}  & 81.5~/~\textbf{93.0} & 79.0 & 82.0  & 80.5  & 66.3 &  \textbf{821}   &  \textbf{+11.6} \\
\bottomrule
\end{tabular}
\end{adjustbox}
\end{table*}

\begin{table*}[h!]
\caption{\footnotesize
Evaluation of 34B models on text benchmarks: MMLU, GSM8K, MATH and HumanEval.
}
\vspace{-.2em}
\centering
\begin{adjustbox}{width={\textwidth}}
\label{tab:34b_text-only}
\begin{tabular}{lccccccc}
\toprule
\multirow{2}{4em}{Tasks}   & Backbone  & \multirow{2}{4em}{MMLU}  & \multirow{2}{4em}{GSM8K} & \multirow{2}{4em}{MATH}  &  \multirow{2}{4em}{HumanEval}  &  Avg.   &  Text-only  \\
& LLM   & & & & & Accuracy & Avg. 4 \\
\midrule
{\color{teal}\emph{(a)}}~Nous-Hermes-2-Yi-34B~\citep{Nous-Hermes-2-Yi-34B} & N/A & 75.5 & 78.6 & 21.8 & 43.3 &  54.8 & N/A \\
\midrule  
NVLM-D 34B &  {\color{teal}\emph{(a)}}  & 73.4 & 82.3 & 47.8 & 62.8 & 66.6 &  \textbf{+11.8} \\ 
NVLM-X 34B &  {\color{teal}\emph{(a)}}  & 73.2  & 82.2 & 46.4 & 62.2 & 66.0 & \textbf{+11.2} \\ 
NVLM-H 34B &  {\color{teal}\emph{(a)}}  & 73.4 & 82.7 & 46.3 & 63.4 & 66.4 & \textbf{+11.6}
\\ 
\bottomrule
\end{tabular}
\end{adjustbox}
\end{table*}

\end{document}